\documentclass[default,iicol]{sn-jnl}% Default with double column layout

\jyear{2023}%

\theoremstyle{thmstyleone}%
\theoremstyle{thmstyletwo}%

\theoremstyle{thmstylethree}%

\usepackage{inconsolata}
\usepackage{CJKutf8}
\usepackage{caption}
\usepackage{bm}
\usepackage{amsmath}
\usepackage{amssymb}
\usepackage{color}
\usepackage{graphicx}
\usepackage{subfigure}
\usepackage{ragged2e}
\usepackage{booktabs}
\usepackage{arydshln}
\usepackage{multirow}
\usepackage{amsfonts,amssymb}
\begin{document}
\begin{sloppypar}
\begin{CJK}{UTF8}{gbsn}
\title[Article Title]{OTIEA:Ontology-enhanced Triple Intrinsic-Correlation for Cross-lingual Entity Alignment}

\author{\fnm{Zhishuo} \sur{Zhang}}\email{2110140@tongji.edu.cn}
\author*{\fnm{Chengxiang} \sur{Tan}}\email{jerrytan@tongji.edu.cn}

\author{\fnm{Xueyan} \sur{Zhao}}\email{1710839@tongji.edu.cn}
\author{\fnm{Min} \sur{Yang}}\email{2110136@tongji.edu.cn}
\author{\fnm{Chaoqun} \sur{Jiang}}\email{tommmm@foxmail.com}

\affil{\orgdiv{The Department of Computer science and technology}, \orgname{Tongji University}, \orgaddress{\city{Shanghai}, \country{China}}}


\abstract{
	Cross-lingual and cross-domain knowledge alignment without sufficient external resources is a fundamental and crucial task for fusing irregular data. As the element-wise fusion process aiming to discover equivalent objects from different knowledge graphs (KGs), entity alignment (EA) has been attracting great interest from industry and academic research recent years. Most of existing EA methods usually explore the correlation between entities and relations through neighbor nodes, structural information and external resources. However, the complex intrinsic interactions among triple elements and role information are rarely modeled in these methods, which may lead to the inadequate illustration for triple. In addition, external resources are usually unavailable in some scenarios especially cross-lingual and cross-domain applications, which reflects the little scalability of these methods. To tackle the above insufficiency, a novel universal EA framework (OTIEA) based on ontology pair and role enhancement mechanism via triple-aware attention is proposed in this paper without introducing external resources. Specifically, an ontology-enhanced triple encoder is designed via mining intrinsic correlations and ontology pair information instead of independent elements. In addition, the EA-oriented representations can be obtained in triple-aware entity decoder by fusing role diversity. Finally, a bidirectional iterative alignment strategy is deployed to expand seed entity pairs. The experimental results on three real-world datasets show that our framework achieves a competitive performance compared with baselines.
	
	%As a fundamental and crucial task in natural language processing(NLP), entity alignment(EA) aims to discover the entities referring to the same real-world object from different knowledge graphs(KGs). 
	%Most of existing EA methods usually generate entity or relation representation via neighbor nodes, structural information and external resources.
}




\maketitle
\section{Introduction}
Knowledge graphs(KGs) have been successfully applied in many real-world scenarios in the past decade. Some cross-lingual KGs such as DBpedia \citep{bizer_dbpedia_2009}, YAGO \citep{suchanek_yago_2008} and ConceptNet \citep{speer_conceptnet_2018} have been built by some commercial companies for recommendation system \citep{recommand}, information retrieve \citep{question} and other knowledge-based downstream tasks \citep{medical1}. Establishing the connectivity among different KGs to fuse useful information has been a hot-spot in recent years. However, heterogeneity and redundancy among domain-specific and language-specific KGs make great limitations on the knowledge integration. Entity alignment(EA) is a fundamental and critical task for connecting different KGs while filtering redundant information from different domains and languages by discovering equivalent entities representing the same real-world object.

Embedding-based methods have been the mainstream for EA \citep{wang_cross-lingual_2018, mao_relational_2020, peng_embedding-based_2020, zhang_adaptive_2021}, which maps entities and relations of KGs into different latent low-dimensional vector spaces via different techniques. These methods can be roughly classified into two categories according to motivations: \textbf{Trans}-based methods and \textbf{GNNs}-based methods. The former models internal interaction among triple elements as explicit translation process as TransE \citep{bordes_translating_2013}. And the latter tries to aggregate information transmission of neighbor nodes to enrich semantic representation via Graph Convolution Networks(GCNs) and Graph Attention Networks(GATs). Both methods have achieved significant improvement for EA, however there are still some important issues that have not been fully considered which may be one of the causes affecting the performance of previous methods.

(1) Some researchers proposed that it is insufficient to rely only on the internal knowledge to represent relational triple in KGs and some external resources of entities can make a significant supplement to enrich entity representation. Previous works \citep{trisedya_entity_2019, li_unsupervised_2019, JAPE, yang_cotsae_2020, zhu_cross-lingual_2022, gao_mhgcn_2022, tang_bert-intbert-based_2020, yang2019aligning, zhang_improving_2022-1, tam_entity_2022} have achieved considerable results attempting to use the attribute triple, entity description and other non-text information from external resources. However, regardless of superior performance, these methods are non-generic especially in some low-resource scenarios, in which the external resources are unavailable or untrustworthy due to the domain-specific uniqueness of knowledge.

(2) The triple is indivisible and the elements are interdependent. For a specific triple $(h,r,t)$, the head entity $h$, relation $r$ and the tail entity $t$ have a natural order in external form which reflect not only the shallow semantic combination but also the ensemble triple representation. However, most of the existing GNNs-based methods \citep{gao_mhgcn_2022, dual_gated, zhu_relation-aware_2020, wu_relation-aware_2019, zhu_raga_2021} embed triple elements separately losing potential semantic and ignoring the role difference of entities in KGs. In addition, the intrinsic correlation among triple elements is complex and indescribable, which cannot be fully expressed as Trans-based methods \citep{transh2014, lin_learning_2015, sun_transedge_2019} trying to model the simple mapping process among triple elements.

\begin{figure}[t]
	\centering
	\subfigure[\textit{\textbf{Primary KG}}]{
		\begin{minipage}{0.45\textwidth}
			\centering
			\includegraphics[width=\textwidth]{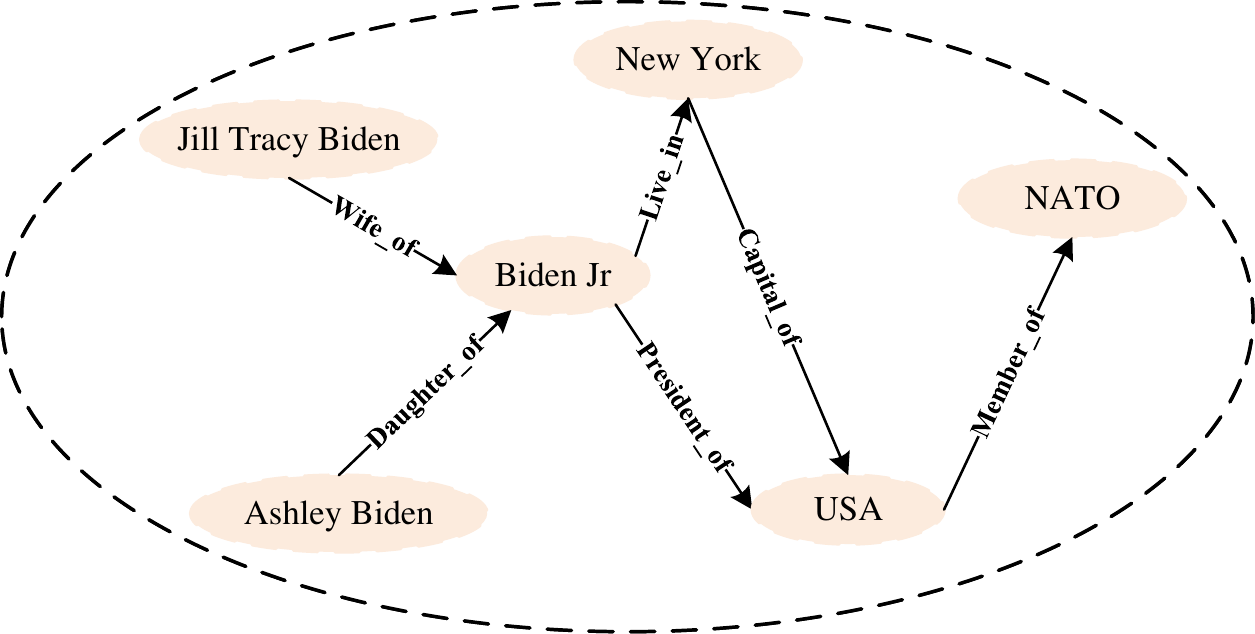}
		\end{minipage}
	}
\subfigure[\textit{\textbf{Ontology KG}}]{
	\begin{minipage}{0.45\textwidth}
		\centering
		\includegraphics[width=\textwidth]{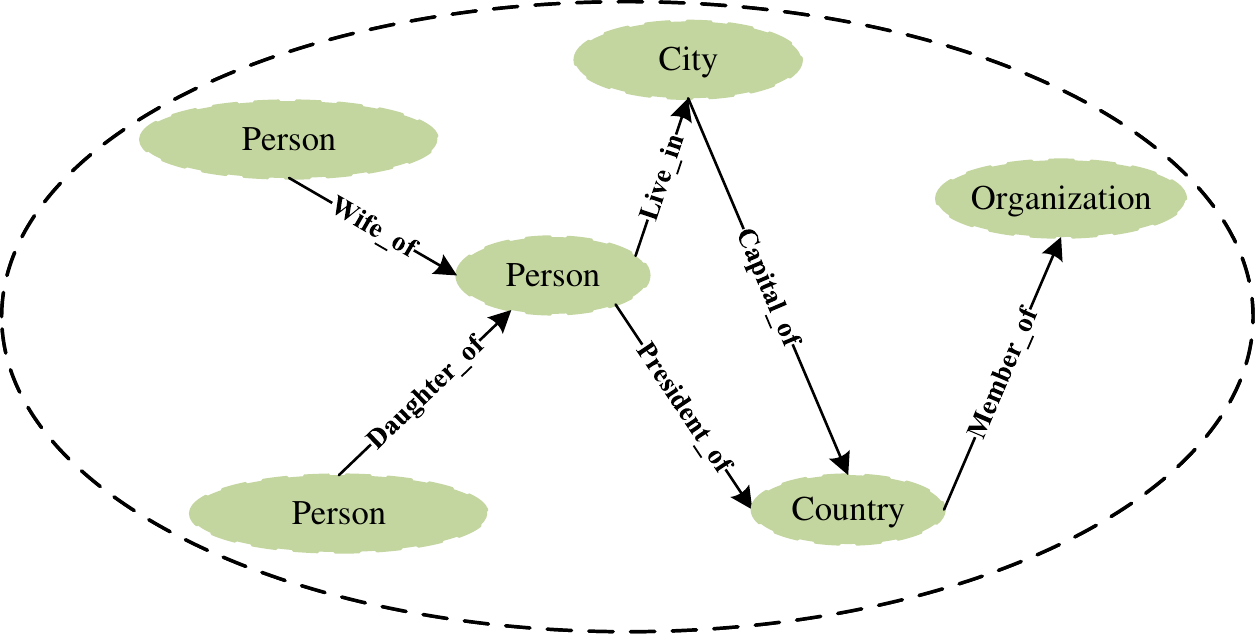}
	\end{minipage}
}
	\caption{An example of the primary KG and its corresponding ontology KG}
	\label{FIG:1}
\end{figure}

(3) The pair-wise ontology is a significant auxiliary resource to enhance semantic triple representation. In our intuitive perspective, the ontology information is the most shallow and simple expression of entities, which contains the primary semantic and type information being ignored by most of existing methods. Furthermore, the ontology pair can provide a more comprehensive semantic representation as a whole, while the ontology of a single entity makes no sense for a triple. As the concrete illustration of primary KG and corresponding ontology KG in Figure \ref{FIG:1}, for the triple \textit{\textbf{(Ashley Biden, Daughter\_of, Biden Jr)}}, its corresponding ontology triple \textit{\textbf{(Person, Daughter\_of, Person)}} means that there is a specific relation type \textit{\textbf{Daughter\_of}} between two persons. It is intuitive that the two \textit{\textbf{Person}} entities should be treat as a whole for the specific triple rather than two separated elements which is meaningless. However, the pair-wise features are rarely contained in previous related works.

Taking the above insight into account, we propose OTIEA -- \textbf{O}ntology-Enhanced \textbf{T}riple \textbf{I}ntrinsic-Correlation for Cross-lingual \textbf{E}ntity \textbf{A}lignment in this paper with the intuitive perspective that triple is indivisible whether in semantic or ontology space, which is capable of fully modeling triple intrinsic correlation and ontology pair enhancement. OTIEA designs a three-stage attention mechanism to mine latent semantic hidden in the process of interaction among elements to build semantic triple representation. And an ontology enhancement method is introduced to fuse ontology pair information. Further, a triple-aware entity decoder is constructed to decode ensemble triple into role-aware entity for EA. Experimental results on three cross-lingual KGs in DBP15K prove that OTIEA achieves competitive peformance compared with baselines. The source code is available in github\footnote{\url{https://github.com/CodesForNlp/OTIEA}}.

The main contributions of the paper are as follows:

$\bullet$ An encoder-decoder architecture is firstly designed and applied into EA task in this paper, which provides a novel perspective to encode ensemble triple.

$\bullet$ A three-stage interaction attention mechanism and a ontology pair enhancement method are designed in encoding process for catching the complex intrinsic correlation in triple while making a supplement to semantic triple via ontology information. And a triple-aware entity decode strategy is introduced in decoder considering the entity role features.

$\bullet$ Extensive experiments conducted on three public datasets demonstrate that OTIEA achieves competitive results compared with state-of-the-art baseline methods.

\section{Related Work}\label{sec2}
Recently, embedding-based representation learning has become the hot-spot both in industrial application and academic research. In this section, we give a detailed illustration of existing embedding-based methods for EA that can be roughly divided into two categories according to representation level:

\textit{\textbf{Trans-based Methods:}} TransE \citep{bordes_translating_2013} is the first paradigm of Trans-based representation learning, which models the triple semantic as a translation process from head entity $\boldsymbol{h}$ to tail entity $\boldsymbol{t}$ across relation $\boldsymbol{r}$ as $\boldsymbol{h\!+\!r\! \approx \!t}$. However, due to the insufficient capacity of representation, the overlapping relations can not be effectively solved. Thus, some researchers explore available solutions to tackle overlapping problem based on TransE and other Trans models such as TransH \citep{transh2014}, TransR \citep{lin_learning_2015} and TransEdge \cite{sun_transedge_2019} via different mapping mechanisms of entity and relation. TransR maps entity to its correlated specific-relation latent space, which tackles the limitations of TransE in a certain degree. TransEdge contextualizes relation representations in terms of specific head-tail entity pairs based on TransE. MTransE \cite{chen_multilingual_2017} obtains entity and relation representations in different KGs respectively, and constructs mapping transformation on two KGs for entity alignment. BootEA \cite{sun_bootstrapping_2018} adopts bootstrapping strategy to extend aligned seeds. RpAlign \cite{huang_cross-knowledge-graph_2022} transforms the entity alignment task to the KG completion task and proposes a new 'anchor' relation for aligned entities replacing the entity vector distance for measuring. JAPE \cite{JAPE} incorporates relational triple, structural information and attributes to optimize the representations of entities and relations. IPTransE \cite{zhu2017iterative} maps embeddings of two KGs into a unified latent vector space, and employs iterative alignment strategy to enhance entity representation. These Trans-based models attempt to establish semantic associations between entity and relation in different ways, but the inherent structural information is ignored and the intrinsic correlation among triple components may not be simply characterized by the translation process from head entity to tail entity, which is differently modeled via three-stage internal interaction attention mechanism verified valid in our work.

\textit{\textbf{GNNs-based Methods:}} GNNs are capable to capture the structural information and propagate neighbor messages, which are important for knowledge representation in KGs. Therefore, many efforts have been made to explore the rich information internal KGs via GNNs recently. GCN-Align \cite{wang_cross-lingual_2018} is the first work to utilize structure information to enhance entity representation via GCNs. MRAEA \cite{mao_mraea_2020} models entity embedding by attending over the node's incoming and outgoing neighbours and its connected relations, it proposes a bi-directional iterative strategy for enhancing representation. RREA \cite{mao_relational_2020} leverages relational reflection transformation to obtain relational specific embedding for each entity. MCEA \cite{qi_multiscale_2022-5} utilizes only structural information for entity alignment in which the convolution region of long-tail entities is extended for capability improvement of graph network. NAEA \cite{zhu_neighborhood-aware_2019} incorporates neighborhood subgraph-level information of entities and utilizes neighborhood-aware attention representation. RDGCN \cite{wu_relation-aware_2019} utilizes relation by attentive interaction between relational KGs and dual relation counterpart. NMN \cite{wu_neighborhood_2020} designs a neighborhood matching mechanism to construct matching-oriented entity representations via graph sampling method. KAGNN \cite{huang_multi-view_2022} extends GNNs to a multi-view version and carefully incorporates both knowledge facts and neighboring structures, and proposes a knowledge-aware attention mechanism to preserve the original semantic and determine the importance of knowledge facts. SHEA \cite{yan_soft-self_2021} takes both relation semantic and edge direction into consideration, aggregates information from entity neighbors in different cases via soft-attention and hard-attention mechanisms. RAGA \cite{zhu_raga_2021} adopts self-attention mechanism to spread entity information to relations and then aggregate relation information back to entities for tackling multiple relations in KGs. Existing GNNs-based representation methods effectively improve the performance of entity alignment. However, the specific features inside triple still cannot be reliably modeled and the utilization of ontology and role information of entity pair is insufficient. To tackle these shortcomings, the ontology pair and role diversity are fused into OTIEA through ontology-enhanced triple encoder and triple-aware entity decoder.

\section{Problem Formulation}\label{sec3}
A KG could be formalized as $KG = (E, R, T)$, where $E=\{e_1, e_2, ..., e_n\}$ and $R=\{r_1, r_2, ..., r_k\}$ are the sets of entity and relation respectively, $n$ and $k$ are the numbers of entities and relations in KG respectively, $T\! \subset \!E\! \times \!R\! \times \!E$ is the set of relational triple. Given two cross-lingual KGs, $KG1=(E1, R1, T1)$ and $KG2=(E2, R2, T2)$, the task of entity alignment is defined as discovering entity pairs referring to the same real-world object in $KG1$ and $KG2$ based on a set of seed entity pairs, which is denoted as $S=\{(e_1, e^\prime_1)\lvert e_1\in E1, e^\prime_1 \in E2\}$, where $e_1$ and $e^\prime_1$ are equivalent.
\section{OTIEA Framework}\label{sec4}
\begin{figure*}[htbp]
	\centering
	\includegraphics[scale=.77]{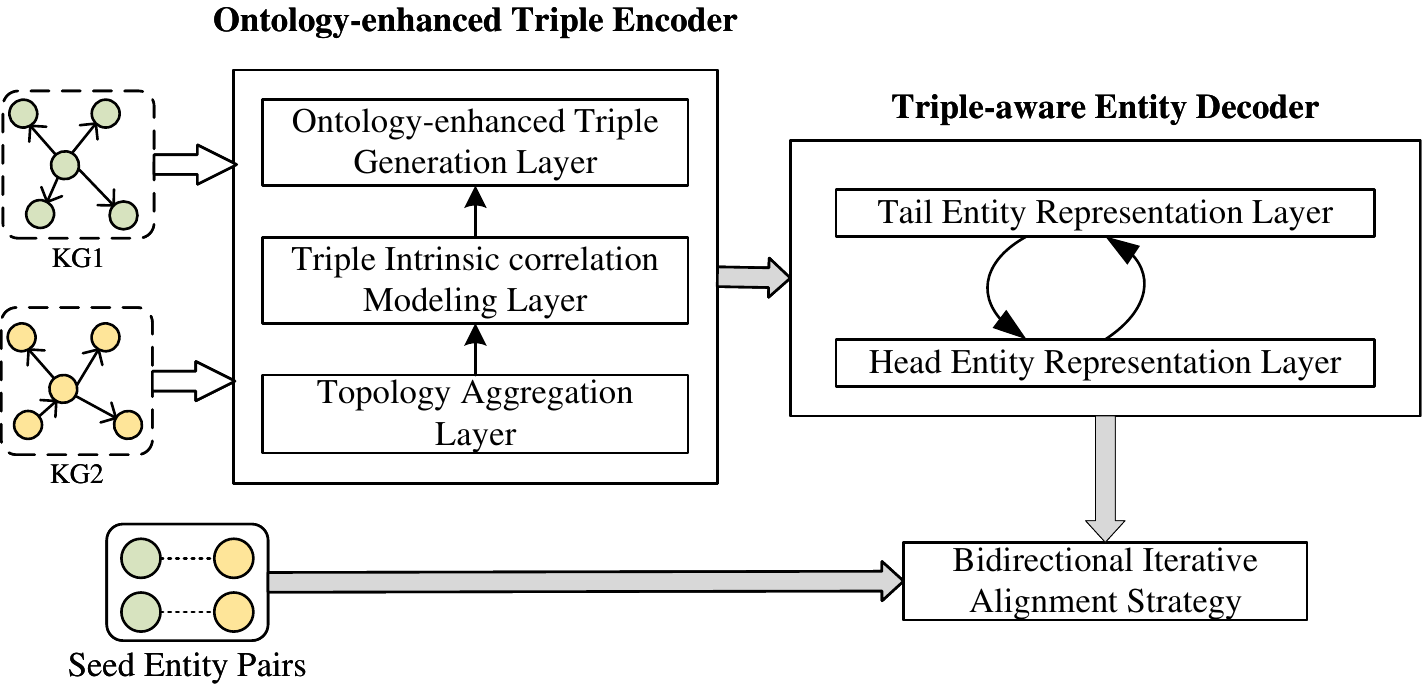}
	\caption{The overall architecture of OTIEA framework, which adopts the encoder-decoder structure. The initialized entities in two KGs are separately fed into Ontology-enhanced Triple Encoder to generate ensemble triple representations via a three-stage interaction attention mechanism and ontology pair enhancement. Then the triple representations are decoded into EA-orient entities representations fusing role features in Triple-aware Entity Decoder. Finally, a bidirectional iterative alignment strategy is deployed to expand training seed pairs.}
	\label{FIG:2}
\end{figure*}

We propose a novel framework OTIEA based on modeling the intrinsic correlation among triple elements and utilizing the importance of ontology pair and role features for cross-lingual EA, which is naturally capable of tackling overlapping relations problem. The overall architecture of OTIEA is illustrated in Fig. \ref{FIG:2}, which can be divided into there parts: Ontology-enhanced Triple Encoder, Triple-aware Entity Decoder and Alignment Strategy. Firstly, two KGs are separately fed into the Ontology-enhanced Triple Encoder, in which structural information is aggregated in GCNs-based Topology Aggregation Layer, and a Triple Intrinsic Correlation Modeling Layer is followed to construct ensemble semantic triple and then an Ontology-enhanced Triple Generation Layer is designed to fuse ontology information to encode fused triple representation. Then a triple-aware entity decoder is exploited to decode fused triple representation into entity representation considering element diversity. Finally, a bidirectional iterative alignment strategy is deployed to automatically enlarge seed entity pairs for semi-supervised training process.

\subsection{Ontology-enhanced Triple Encoder}\label{subsec4}
\subsection*{GCNs-based Topology Aggregation Layer}\label{subsubsec4}
In order to fully consider the abundant semantic and associative information transmission through the topology structure of KGs, we firstly make a combination of original-relation, reverse-relation and self-relation, after which the extended KGs are separately fed into a two-layer GCNs-highway network as previous works \cite{zhu_raga_2021} to embed entity nodes.
The GCNs-based topology aggregation generates entity embedding as below:

\begin{equation}
	\boldsymbol{X}^{l+1}=\operatorname{ReLU}\left(\boldsymbol{\tilde{D}}^{-\frac{1}{2}} \boldsymbol{\tilde{A}} \boldsymbol{\tilde{D}}^{-\frac{1}{2}} \boldsymbol{X}^{l}\right)  \label{eq1}
\end{equation}
where $\!\boldsymbol{\tilde{A}}\!\!=\!\boldsymbol{A}\!+\!\boldsymbol{I}$, $\boldsymbol{A}$ is the adjacency matrix of relation-expanded graph, $I$ is the identity matrix, $\boldsymbol{\tilde{D}}$ is the degree matrix of $\boldsymbol{\tilde{A}}$ and $\!\boldsymbol{X}^{l}\!\!\in\! \mathbb{R}^{n\!\times\!d_{e}}$ denotes the input entity embedding in $l$-th hidden layer, $n$ is the number of entities in a KG, $d_{e}$ is the dimension of entity embedding, $\boldsymbol{X}^{l+1}$ is the output of $l$-th layer.

After that, layer-wise highway network \citep{srivastava_highway_2015} is deployed to preserve original entity information avoiding gradient explosion. The Highway network works as follow：
\begin{equation}
	T\left(\boldsymbol{X}^{l}\right) = \sigma\left(\boldsymbol{X}^{l} \boldsymbol{W}^{l}+\boldsymbol{b}^{l}\right) \label{eq2}
\end{equation}
\begin{equation}
	\boldsymbol{X}^{l+1}\!=\!T\left(\boldsymbol{X}^{l}\right)\! \cdot \! \boldsymbol{X}^{l+1}\!+\!\left(1\!-\!T\!\left(\!\boldsymbol{X}^{l}\!\right)\!\right)\! \cdot \!\boldsymbol{X}^{l} \label{eq3}
\end{equation}
where $T\left(\cdot\right)$ is the transmission factor. $\sigma(\cdot)$ is activation function, $\cdot$ denotes the element-wise multiplication, $\boldsymbol{W}^{l}$ and $\boldsymbol{b}^{l}$ are the weight matrix and bias vector of the input embedding in $l$-th hidden layer.
\subsection*{Triple Intrinsic Correlation Modeling Layer}\label{subsubsec4}
In this layer, we regard the ordered concatenation of any two elements in triple as the latent representation of another element and model intrinsic correlation among elements via attention mechanism to construct the ensemble representation of semantic triple. In this process, for a specific triple $(e_i,r,e_j)$ corresponding to head entity embedding $\boldsymbol{X}_i\! \in \!\mathbb{R}^{d_e}$ and tail entity embedding $\boldsymbol{X}_j\! \in \!\mathbb{R}^{d_e}$, the relation latent relation representation $\boldsymbol{X}^r\! \in \!\mathbb{R}^{d_r}$ can be firstly obtained.
\begin{equation}
	\boldsymbol{X}^r=\sigma\left(\left(\boldsymbol{X}_{i} \| \boldsymbol{X}_{j}\right) \boldsymbol{W}_{s r}+\boldsymbol{b}_{s r}\right)
\end{equation}
where $d_e$ and $d_r$ are the dimensions of entity and relation respectively, $\sigma$ denotes the activation function, $\boldsymbol{W}_{s r}\! \in \!\mathbb{R}^{2d_e\! \times \!d_r}$ and $\boldsymbol{b}_{s r}\! \in \!\mathbb{R}^{d_r}$ are trainable parameters, $\|$ is the concatenation operation.

Then a three-stage internal interaction attention mechanism is designed to reflect the influence on a element from other two elements in a triple. Due to the same computing process, we take the internal interaction attention of head entity as an example:
Firstly, we regard the ordered concatenation of relation $\boldsymbol{X}^r$ and tail entity $\boldsymbol{X}_j$ as the latent representation of head entity $\boldsymbol{X}_i$, then the internal interaction attention is computed as:
\begin{equation}
	{\left.c_{irj}=a^{T}\left(\boldsymbol{X}_{i} \boldsymbol{W}_{h} \|\left(\boldsymbol{X}^{r} \| \boldsymbol{X}_{j}\right) \boldsymbol{W}_{h T}\right)\right)}
\end{equation}

\begin{equation}
	\begin{split}
		\alpha_{it}&=\operatorname{Softmax} \left(c_{irj}\right)\\
		&=\frac{\exp \left(\operatorname{LeakyRelu}\left(c_{i r j}\right)\right)}{\sum\limits_{\left(e_{i^{\prime}}, r, e_{j^{\prime}}\right) \in T_{r}} \exp \left(\operatorname{LeakyRelu}\left(c_{i^{\prime} r j^{\prime}}\right)\right)}
	\end{split}
\end{equation}

\begin{equation}
	\boldsymbol{X}_{h i}=\operatorname{Relu}\left(\sum_{e_{i} \in h_{r}}\left(\alpha_{it} \boldsymbol{X}_{i} \boldsymbol{W}_{h}\right)\right)
\end{equation}

where $\boldsymbol{W}_{h}\! \in \!\mathbb{R}^{d_e\! \times \!d_r}$ and $\boldsymbol{W}_{h T}\! \in \!\mathbb{R}^{2d_e\! \times \!d_r}$ are the trainable parameters of head entity and triple intrinsic correlation. $T_{r}$ denotes the subset of triple set $T$ with a specific relation $r$. $h_r$ is the set of head entity from $T_{r}$. $\boldsymbol{X}_{h i}$ is the latent semantic representation of head entity $e_i$.

In the same way, the latent semantic representation of tail entity $e_j$ and relation $r$ can be respectively obtained as $\boldsymbol{X}_{t j}$ and $\boldsymbol{X}^{tr}$. Then the latent semantic representation of triple $(e_i,r,e_j)$ is depicted as $\boldsymbol{S}^c_{irj}=\boldsymbol{X}_{h i}+\boldsymbol{X}^{tr}+\boldsymbol{X}_{t j}$.

Finally, the overlapping relational triple are taken into account, we generate the global feature of triple relation by averaging the concatenation of head entities and tail entities corresponding to a same relation. And the global triple representation which preserves primary feature can be obtained.
\begin{equation}
	\boldsymbol{X}^{r}_g=\left(\frac{1}{\lvert T_{r} \rvert} \sum\limits_{\left(e_{i}, r, e_{j}\right) \in T_{r}}\left(\boldsymbol{X}_{i} \| \boldsymbol{X}_{j}\right)\right)\boldsymbol{W}_{r g}+\boldsymbol{b}_{r g}
\end{equation}

\begin{equation}
	{\boldsymbol{S}^g_{irj}=\sigma\left(\left(\boldsymbol{X}_{i}\left\|\boldsymbol{X}^{r}_g\right\| \boldsymbol{X}_{j}\right) \boldsymbol{W}_{s p}+\boldsymbol{b}_{s p}\right)}
\end{equation}
Where $\boldsymbol{W}_{r g}\! \in \!\mathbb{R}^{2d_e\! \times \!d_r}$, $\boldsymbol{W}_{s p}\! \in \!\mathbb{R}^{(2d_e+d_r)\! \times \!d_r}$, $\boldsymbol{b}_{r g}\! \in \!\mathbb{R}^{d_r}$ and $\boldsymbol{b}_{s p}\! \in \!\mathbb{R}^{d_r}$ are trainable parameters.

After that, the ensemble semantic triple is represented as $\boldsymbol{S}_{irj}=\boldsymbol{S}^g_{irj}+\boldsymbol{S}^c_{irj}$.

\subsection*{Ontology-enhanced Triple Generation Layer}\label{subsec4}
In this layer, the ontology information is fused into triple semantic to form enhanced triple representation. Next, we try to introduce the enhancement process in detail.

Firstly, we transform entity representation from semantic space to latent ontology space regarding semantic and ontology as different modals.
\begin{equation}
	{\boldsymbol{X}^{o}=\tanh \left(\boldsymbol{W}_{s2o} \boldsymbol{X}+\boldsymbol{b}_{s2o}\right)}
\end{equation}
where $\boldsymbol{X}^{o}$ denotes the ontology representation of entities, $\boldsymbol{W}_{s2o}\! \in \!\mathbb{R}^{d_e\! \times \!d_o}$ and $\boldsymbol{b}_{s2o}\! \in \!\mathbb{R}^{d_o}$, $d_o$ is the dimension of ontology representation.

Then, due to it is meaningless to be handled for a single entity ontology for a specific triple, thus entity pair is treated as a whole in ontology space. What is more, the entity pair with same ontology representations may related to more than one relation, therefore, the global ontology relation representation is obtained as:
\begin{equation}
	\boldsymbol{X}^{or}_g\!=\!\frac{1}{\lvert T_{r} \rvert}\! \sum\limits_{\left(e_{i}, r, e_{j}\right) \in T_{r}}\!\left(\boldsymbol{X}^{o}_{i} \| \boldsymbol{X}^{o}_{j}\right)  \label{eq11}
\end{equation}
\begin{equation}
	\boldsymbol{X}^{or}\!=\!(\boldsymbol{X}^{or}_g)\boldsymbol{W}^{o}_{r g}\!+\!\boldsymbol{b}^{o}_{r g}  \label{eq12}
\end{equation}
where $\boldsymbol{W}^{o}_{r g}\! \in \!\mathbb{R}^{d_o\! \times \!d_r}$ and $\boldsymbol{b}^{o}_{r g}\! \in \!\mathbb{R}^{d_r}$ are trainable parameters. Then the ontology triple can be formed as:
\begin{equation}
	{\boldsymbol{O}_{i r j}=\sigma\left(\left(\boldsymbol{X}^{o}_{ei}\left\|\boldsymbol{X}^{or}\right\| \boldsymbol{X}^{o}_{ej}\right) \boldsymbol{W}^{o}_{t}+\boldsymbol{b}^{o}_{t}\right)}
\end{equation}
where $\boldsymbol{W}^{o}_{t}\! \in \!\mathbb{R}^{(2d_o+d_r)\! \times \!d_r}$ and $\boldsymbol{b}^{o}_{t}\! \in \!\mathbb{R}^{d_r}$ are trainable parameters.

We believe that the two modalities of semantic and ontology representations are mutually reinforced, a co-attention mechanism between the two modals is designed to get semantic and ontology fused ensemble triple representation. The co-attention mechanism is described as:

\begin{eqnarray}
	\!\alpha_{t r}\!\!=\!\frac{\exp \left(\!\operatorname{LReLU}\!\!\left(\!a^{T}\!\left(\!\boldsymbol{S}_{irj} \| \boldsymbol{{O}}_{irj}
	\!\right)\right)\right)\!}{\!\sum\limits_{T_r}\! \exp \left(\!\operatorname{LReLU}\!\!\left(\!a^{T}\!\left(\!\boldsymbol{S}_{i^{\prime}rj^{\prime}} \| \boldsymbol{{O}}_{i^{\prime}rj^{\prime}} \!\right)\right)\right)\!} \label{eq18}
\end{eqnarray}
\begin{equation}
	\boldsymbol{\bar{O}}_{r}=\operatorname{ReLU}\left(\sum_{(e_i,r,e_j) \in T_{r}}\left(\alpha_{t r} \boldsymbol{S}_{irj}\right)\right) \label{eq19}
\end{equation}
The enhanced ontology triple representation $\boldsymbol{\bar{O}}_{r}$ is obtained by above computing process. As the same, the enhanced semantic triple representation $\boldsymbol{\bar{S}}_{r}$ can be generated.

Finally, preserving primary ontology features, we form the ontology-enhanced ensemble triple representation $\boldsymbol{T}_{irj}\! \in \!\mathbb{R}^{2d_o\!+\!d_r}$.
\begin{equation}
	\boldsymbol{T}^{\prime}_{irj}\!=\!\boldsymbol{S}_{irj}\!+\!\boldsymbol{\bar{S}}_{r}\!+\!\boldsymbol{\bar{O}}_{r}\!+\!\boldsymbol{{O}}_{irj}
\end{equation}
\begin{equation}
	\boldsymbol{T}_{irj} = \boldsymbol{T}^{\prime}_{irj} \| \boldsymbol{X}^{or}_g \label{eq20}
\end{equation}

\subsection{Triple-aware Entity Decoder}\label{subsec4}
\subsubsection*{Cycle co-Enhanced Mechanism}\label{subsubsec4}
The triple-aware entity decoder is designed to distinguish the role diversity in which a entity pair cycle co-enhanced mechanism is deployed to improve entity representation ability which is verified to be valid in Section \ref{sec6}.

The entity pair cycle co-enhanced mechanism is composed of three triple-aware attention layers corresponding to different entity roles: two head-aware entity representation layers and one tail-aware entity representation layer. The organization form of a head-aware layer, a tail-ware layer and another head-aware layer guarantee the full use of entity role information. Considering the computation similarity of three triple-aware attention layers, we just formulated the head-aware entity representation process here as follows:
\begin{eqnarray}
	\alpha_{rj}\!=\! \frac{\!\exp \!\left(\!\operatorname{LReLU}\!\!\left(\!a^{T}\!\left(\!\boldsymbol{T}_{irj}\boldsymbol{W}^{h} \| \boldsymbol{X}_{i}\!\right)\right)\right)\!}{\!\sum\limits_{\!\left(\!e_i,r^{\prime},e_j^{\prime}\!\right)\! \in T}\!\!\exp \!\left(\!\operatorname{LReLU}\!\!\left(\!a^{T}\!\left(\!\boldsymbol{T}_{ir^{\prime}j^{\prime}}\boldsymbol{W}^{h} \| \boldsymbol{X}_{i}\!\right)\right)\right)\!} \label{eq21}
\end{eqnarray}
\begin{equation}
	\boldsymbol{X}_{i}\!=\!\boldsymbol{X}_{i}\!+\!\operatorname{ReLU}\!\left(\!\sum_{\left(e_i,r,e_j\right) \in T}\left(\alpha_{rj} \boldsymbol{T}_{irj} \boldsymbol{W}^{h}\right)\!\right)\! \label{eq22}
\end{equation}
where $\boldsymbol{W}^{h}\!\in \!\mathbb{R}^{(d_r\!+\!2d_o)\! \times \!d_e}$ is the trainable weight parameters. And then the head-ware entity representation is put into the computation of tail-aware entity representation. Then the ontology-enhanced entity representation can be obtained, which is put into a primary GAT layer to re-aggregate neighbor information to generate final entity representation $\boldsymbol{X}^{f}=\{\boldsymbol{x}^f_1, \boldsymbol{x}^f_2, ..., \boldsymbol{x}^f_n\}$.

\subsection{Iterative Alignment Strategy}\label{subsec4}
After obtaining the final entity representation, a iterative alignment strategy \cite{mao_mraea_2020} is deployed to execute semi-supervised training process. And the margin-based loss function with L1 norm distance is defined as follows:

\begin{equation}
	{dis}\left(e_{i}, e_{j}\right)=\left\|\boldsymbol{x}_{i}^{f}-\boldsymbol{x}_{j}^{f}\right\|_{1} \label{eq25}
\end{equation}

\begin{eqnarray}
	%\begin{aligned}
	L\!= \!\sum_{\!\left(\!e_{i}, e_{j}\!\right)\! \in S}\max\!\left(\!{dis}\!\left(\!e_{i}, e_{j}\!\right)\!\!-\!{dis}\!\left(\!e_{i}^{\prime}, e_{j}^{\prime}\!\right)\!\!+\!\lambda, 0\!\right)\! \label{eq26}
\end{eqnarray}
%\end{small}

where $(e_{i}, e_{j})$ is the pre-aligned entity pair in training set $S$, $(e_i^{\prime}, e_j^{\prime})$ is the negative sample generated by randomly replacing $e_{i}$ or $e_{j}$ with their $k$-nearest neighbors, $\lambda$ is the margin hyper-parameter. In the process of bidirectional semi-supervised training, the minimum entities distances from source KG to target KG and from target KG to source KG are computed, respectively. And only the entity pairs with both the bidirectional minimum distances are added to training sets, which effectively alleviate the wrong aligning entity pairs introduced in the process of semi-supervised training.

The overall process of OTIEA framework is described as Algorithm \ref{alg1}.
\begin{algorithm}[H]
	\caption{The Overall Process of OTIEA Framework}\label{alg_alg1}
	\begin{algorithmic}[1]
		\Require Source {$KG_s$}, Target {$KG_t$}
		\State Initialize $E_s$ and $E_t$ via Glove
		\For {$epoch\ i=1 \rightarrow E$}
			\State {Aggregate topological structures of $E_s$ and $E_t$ by Equations \eqref{eq1}--\eqref{eq3}}
			\State Generate Semantic Triple $S_s$ and $S_t$
			\State Obtain Ensemble Triple $T_s$ and $T_t$ by Equation \eqref{eq20}
			\State Generate EA-orient entity representation $E_s$ and $E_t$ by Equations \eqref{eq21}--\eqref{eq22}
			\State Training with Equations \eqref{eq25}--\eqref{eq26}
			\If{$n\%5 == 0$}
				\State Expanding Training Sets $S$
			\EndIf
		\EndFor
	\end{algorithmic}
	\label{alg1}
\end{algorithm}

\section{Experimental Setup}\label{sec5}
\subsection{DataSets}\label{subsec5}
To make fair comparisons with baselines, we evaluate OTIEA framework on the simplified DBP15K, which is a public data set widely used by almost all related methods for cross-lingual EA task and is described in Table \ref{lab1}. It is noted that the difference between simplified version and primary version is that the isolated entities are removed which provide little assistance for most of EA methods relying on message propagation through the structure of KGs.

As the Table \ref{lab1} shows, the simplified DBP15K contains three cross-lingual sub-datasets in Chinese, English and French: ZH\_EN, JA\_EN and FR\_EN. All the numbers of linking triples in three sub-datasets are 15000, and the numbers of entities are all 19k, the numbers of relations are at the same level, the number of triples in FR\_EN is slightly more than that in ZH\_EN and JA\_EN.
\begin{table}[h]
	\caption{Statistical data of the Simplified DBP15K.} \label{lab1}
	\resizebox{\linewidth}{!}{
		\begin{tabular}{cccccc}
			\toprule%
			{DBP15K} &{} & {Entities} & {Relations} & {Rel Triples} & {Links}\\
			\midrule
			\multirow{2}{*}{ZH-EN}
			&{ZH} & {19388} & {1700} & {70414}& \multirow{2}{*}{15000}\\
			&{EN} & {19572} & {1322} & {95142} &\\
			\midrule
			\multirow{2}{*}{JA-EN} 
			&{JA} & {19814} & {1298} & {77214} &\multirow{2}{*}{15000}\\
			&{EN} & {19780} & {1152} & {93484} &\\
			\midrule
			\multirow{2}{*}{FR-EN} 
			&{FR} & {19661} & {902} & {105998} &\multirow{2}{*}{15000}\\
			&{EN} & {19993} & {1207} & {115722} &\\
			\bottomrule 
		\end{tabular}
	}
\end{table}
\subsection{Baseline}\label{subsec5}
To evaluate OTIEA from multiple aspects, we choose methods without introducing external resources as baselines, which can be classified into two categories. Further, some related methods with the semi-supervised training technique implemented by us or provided by original papers are also compared to prove model superiority from different levels.

\textbf{Trans-based methods (basic):} MTransE \citep{chen_multilingual_2017}, BootEA \citep{sun_bootstrapping_2018}, TransEdge \citep{sun_transedge_2019}, RpAlign \citep{huang_cross-knowledge-graph_2022}.

\textbf{GNNs-based methods (basic):} GCN-Align \citep{wang_cross-lingual_2018}, MCEA \citep{qi_multiscale_2022-5}. NAEA \citep{zhu_neighborhood-aware_2019}, RDGCN \citep{wu_relation-aware_2019}, NMN \citep{wu_neighborhood_2020}, KAGNN \citep{huang_multi-view_2022}, SHEA \citep{yan_soft-self_2021}, RAGA \citep{zhu_raga_2021}.

\textbf{Semi-supervised methods:} 
MRAEA-semi \citep{mao_mraea_2020},
RREA-semi \citep{mao_relational_2020}, RAGA-semi \citep{zhu_raga_2021}, MCEA-semi \citep{qi_multiscale_2022-5}.

To our best knowledge, RAGA achieves state-of-the-art results exploring the interaction between entities and relations without using external resources.

\subsection{Model Variants}\label{subsec5}
We construct three variants of OTIEA-base for prove the effectiveness of different components or modules.

(1) wo-E: a simplified OTIEA version without global feature of triple in Triple Intrinsic Correlation Modeling Layer.

(2) wo-O: a simplified OTIEA version without Ontology-enhanced Triple Generation Layer.

(3) wo-C: a simplified OTIEA version without Cycle co-Enhancement mechanism in Triple-aware Entity Decoder.

\subsection{Implementation Details}\label{subsec5}
We utilize the initialized entities embedding from RAGA, which firstly translates entity names of French and Japanese into English, and Glove model then is deployed to obtain entities embedding. For the three sub-datasets, we adopt same super-parameters settings: the ontology dimension $d_o$ and relation dimension $d_r$ are 100, the GCN's depth $l$ is 2 and the margin hyper-parameter $\lambda$ in loss function is set to 3.0. And we sample nearest $k=$5 neighbors to construct negative samples. By conventional, the 30\% of seed entity pair are used as train set, and the rest for test. And we apply the Hits@k and MRR for all methods to measure the EA performance, the higher scores of all metrics mean the better performance.

\begin{table*}[!h]
	\caption{Overall performance of entity alignment.\label{tab2}}
	\centering
	\resizebox{\linewidth}{!}{
		\begin{tabular*}{500pt}{@{\extracolsep{\fill}}lccccccccc@{\extracolsep{\fill}}}
			\toprule%
			& \multicolumn{3}{@{}c@{}}{\textbf{ZH-EN}} & \multicolumn{3}{@{}c@{}}{\textbf{JA-EN}} & \multicolumn{3}{@{}c@{}}{\textbf{FR-EN}} \\\cmidrule{2-4}\cmidrule{5-7} \cmidrule{8-10}%
			\textbf{Method}& H@1 &H@10 &MRR& H@1 &H@10 &MRR &H@1& H@10 &MRR\\
			\midrule
			MTransE(2017) &30.8 &61.4 &0.364 &27.8 &57.4 &0.349& 24.4 &55.5&0.335\\
			%JAPE(2017) &41.2 &74.4& 0.490& 36.2 &68.5& 0.476 &32.4& 66.7 &0.430\\ %
			BootEA(2018) &62.9& 84.7 &0.703 &62.2& 85.4& 0.701& 65.3 &87.4 &0.731\\
			TransEdge(2019) &73.5 &91.9 &0.801& 71.9 &93.2 &0.795 &71.0 &94.1& 0.796\\
			RpAlign(2022) &74.8 &88.8 &0.794 &72.9& 89.0& 0.872& 75.2 &89.9& 0.801\\
			\cdashline{1-10}
			GCN-Align(2018) &41.2& 74.4& 0.549 &39.9& 74.4 &0.546& 37.3& 74.5& 0.532\\
			%HMAN(2019) &56.1 &85.9& 0.67 &55.7& 86.0 &0.67 &55.0& 87.6 &0.66\\
			%HGCN(2019) &72.0 &85.7& 0.768 &76.6& 89.7 &0.813 &89.2& 96.1 &0.917\\
			MCEA(2022) &72.4 &93.4 &0.800 &71.9& 94.0& 0.800& 73.9 &95.3& 0.820\\
			NAEA(2019) &65.0 &86.7 &0.720& 64.1 &87.3 &0.718 &67.3 &89.4& 0.752\\
			RDGCN(2019) &70.8& 84.6 &0.751 &76.7& 89.5& 0.812& 88.6& 95.7& 0.908\\
			%MHGCN(2022) &73.2 &86.7 &0.793 &76.7& 89.1& 0.821& 86.4 &92.3& 0.898\\
			NMN(2020) &73.3& 86.9 &0.781 &78.5& 91.2 &0.827 &90.2 &96.7 &0.924\\
			KAGNN(2022) &73.6 &87.3 &0.786 &79.4& 91.1& 0.837& 92.0 &97.6& 0.941\\
			%MRGA(2021) &75.5& 90.5 &0.783 &73.4& 90.3 &0.771 &75.7 &91.7 &0.791\\
			SHEA(2021) &76.3 &91.4 &0.835 &82.1& 93.8& 0.860& 90.5 &97.0& 0.902\\
			RAGA(2021) &79.8 &93.3 &0.847 &82.9& 95.0& 0.875& 91.4 &98.2& 0.940\\
			\midrule
			%		TTEA-base(wo-O) &77.3 &90.3 &0.819 &81.5& 92.5& 0.855& 92.2 &97.7& 0.943\\   %无拓扑结构
			OTIEA-base(wo-E) &79.6 &93.4 &0.846 &82.9& 95.1& 0.874& \textbf{92.4} &98.5& \textbf{0.947}\\   %无三元组全局特征
			OTIEA-base(wo-O) &78.8 &\textbf{93.7} &0.842 &81.3& 94.9& 0.864& 91.3 &98.3& 0.940\\   %无类型空间加强
			OTIEA-base(wo-C) &79.4 &93.4 &0.845 &81.7& 95.0& 0.866& 91.9 &98.5& 0.944\\   %非头尾循环加强
			%		TTEA-base(wo-S) &79.8 &93.4 &0.848 &83.5& 95.3& 0.878& 92.3 &98.6& 0.947\\   %不保留三元组特异性特征
			\textbf{OTIEA-base(ours)} &\textbf{80.1} &93.6 &\textbf{0.851} &\textbf{83.1}&\textbf{95.4}&\textbf{0.877}&\textbf{92.4}&\textbf{98.6}&\textbf{0.947}\\
			\midrule
			MRAEA-semi(2020) &75.2& 92.3 &0.824 &75.3 &93.3 &0.825 &78.1& 94.7 &0.843\\
			RREA-semi(2020) &80.1&94.8 &0.857 &80.2 &95.2& 0.858&82.7 &96.6& 0.881\\
			MCEA-semi(2022) &81.4 &95.6 &0.867 &80.7& 95.7& 0.864& 84.1 &97.0& 0.891\\
			RAGA-semi(2021) &85.7 &96.0 &0.896 &\textbf{88.9}& 97.1& \textbf{0.920}& 94.0 &98.8& 0.958\\
			\textbf{OTIEA-semi(ours)} &\textbf{86.3} &\textbf{96.2} &\textbf{0.900} &88.6&\textbf{97.3}&\textbf{0.920}&\textbf{94.8}&\textbf{99.0}&\textbf{0.965}\\
			\bottomrule
		\end{tabular*}
	}
\end{table*}

\section{Results and Analysis}\label{sec6}
Table \ref{tab2} shows the overall results of different EA methods on three sub-datasets of simplified DBP15K, which are either implemented with the source code or provided by original papers. The dashed line separates Trans-based methods with GNNs-based methods. And baseline methods, different OTIEA variants and semi-supervised methods are separated by solid lines.

\begin{figure*}[!h]
	\centering
	\subfigure[\textit{\textbf{mode1}}]{
		\begin{minipage}{0.20\textwidth}
			\centering
			\includegraphics[width=\textwidth]{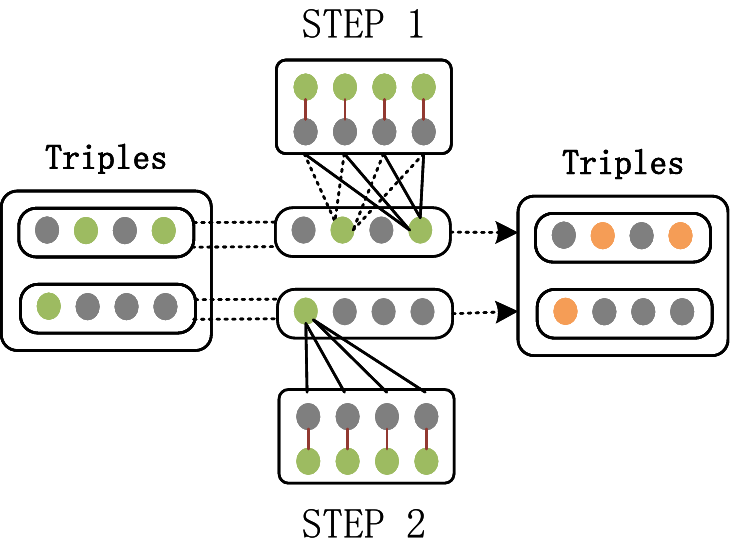}
		\end{minipage}
	}
	\subfigure[\textit{\textbf{mode2}}]{
		\begin{minipage}{0.34\textwidth}
			\centering
			\includegraphics[width=\textwidth]{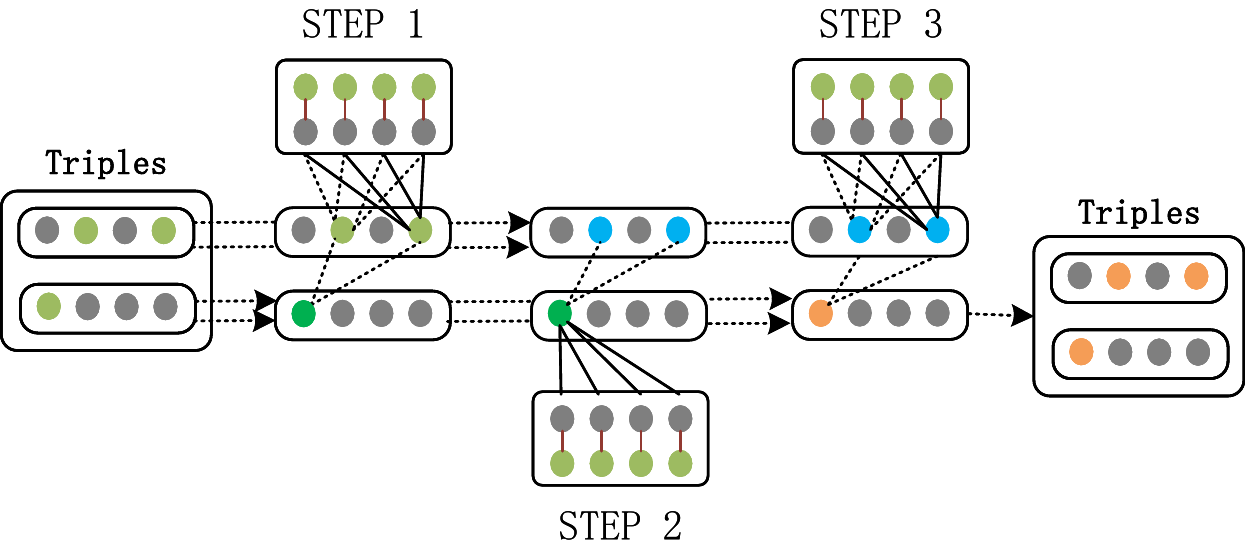}
		\end{minipage}
	}
	\subfigure[\textit{\textbf{mode3}}]{
		\begin{minipage}{0.40\textwidth}
			\centering
			\includegraphics[width=\textwidth]{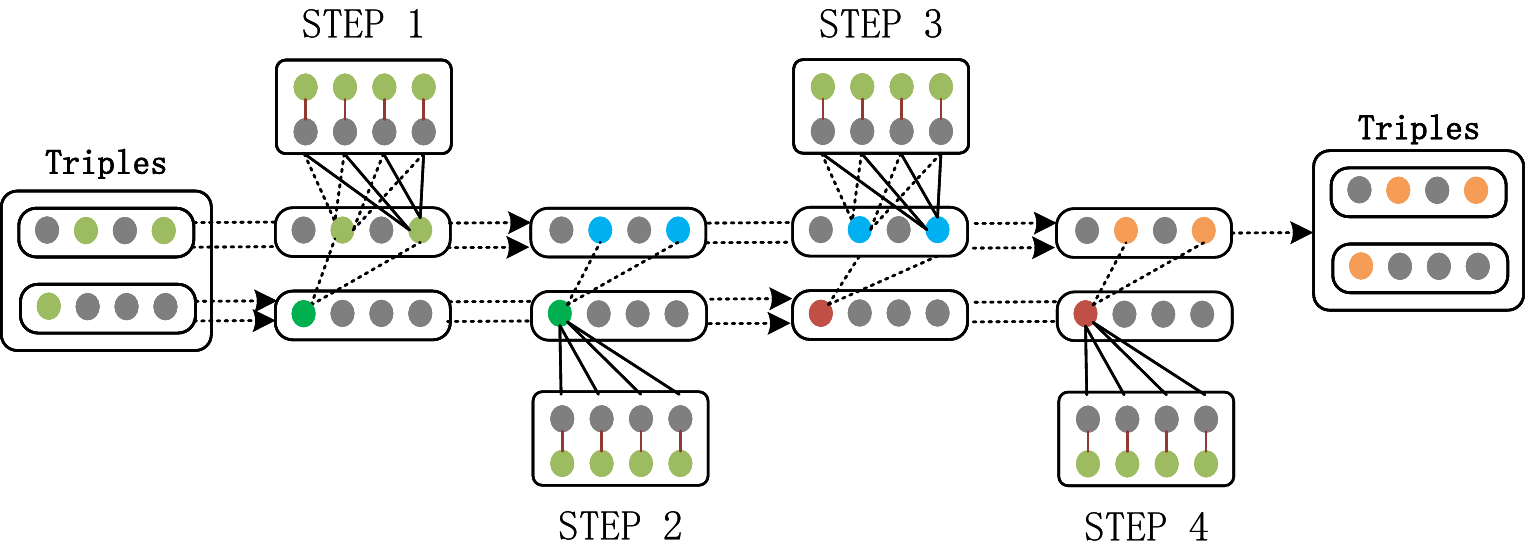}
		\end{minipage}
	}
	\caption{Three different modes of Cycle Co-enhanced Mechanism.}
	\label{FIG:modes}
\end{figure*}

\begin{figure*}[!h]
	\centering
	\subfigure[ZH\_EN]{
		\begin{minipage}{0.31\textwidth}
			\centering
			\includegraphics[width=\textwidth]{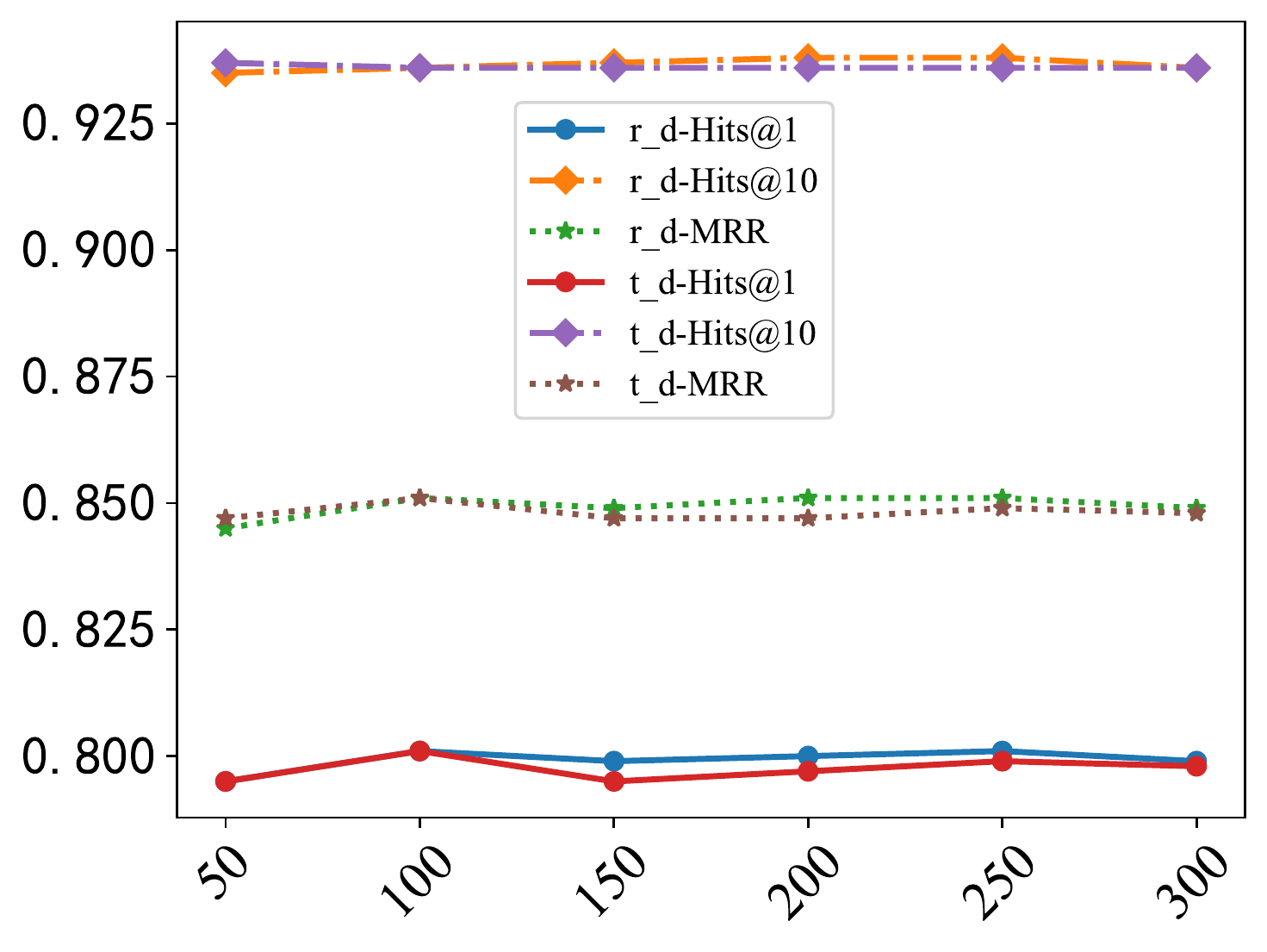}
		\end{minipage}
	}
	\subfigure[JA\_EN]{
		\begin{minipage}{0.31\textwidth}
			\centering
			\includegraphics[width=\textwidth]{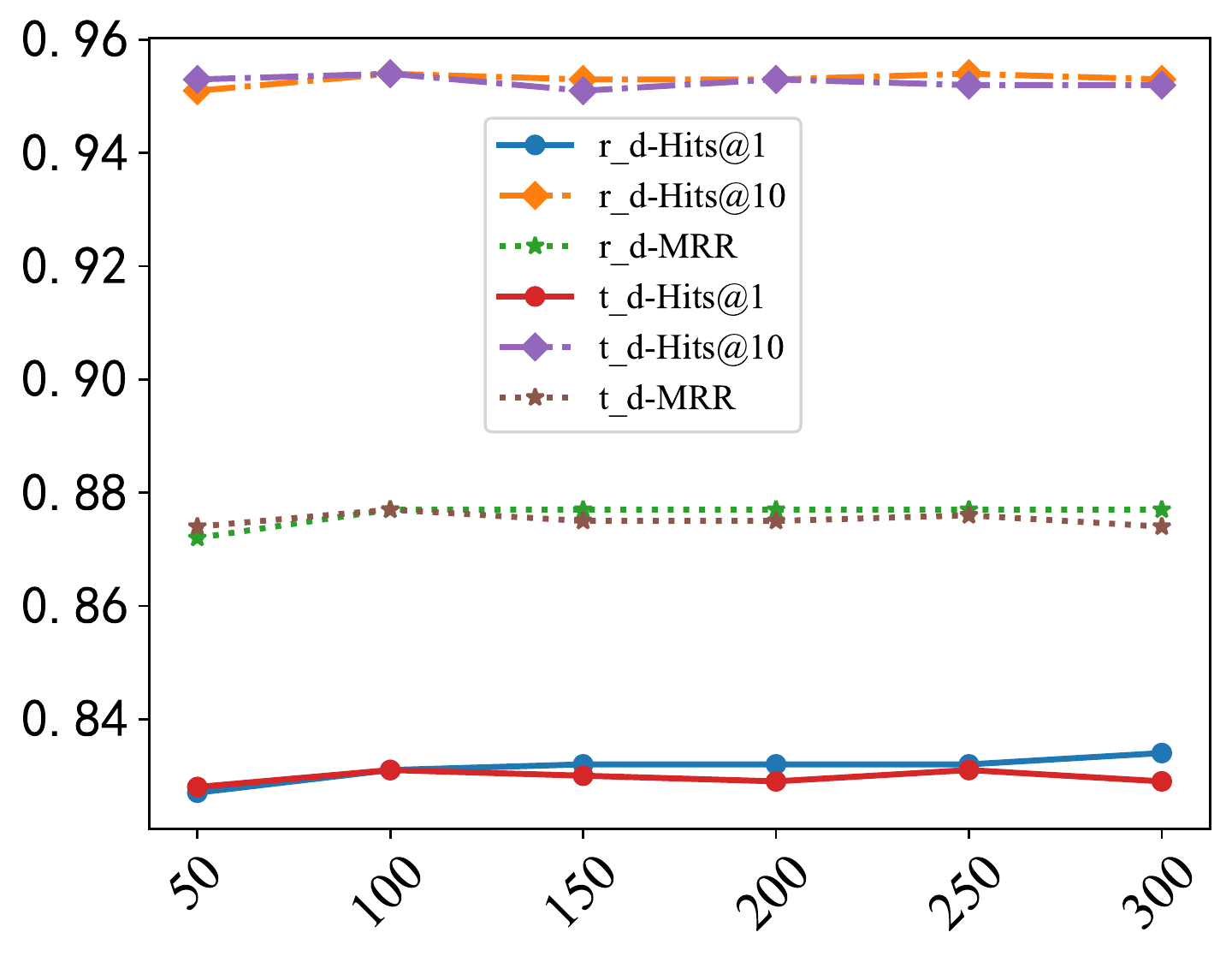}
		\end{minipage}
	}
	\subfigure[FR\_EN]{
		\begin{minipage}{0.31\textwidth}
			\centering
			\includegraphics[width=\textwidth]{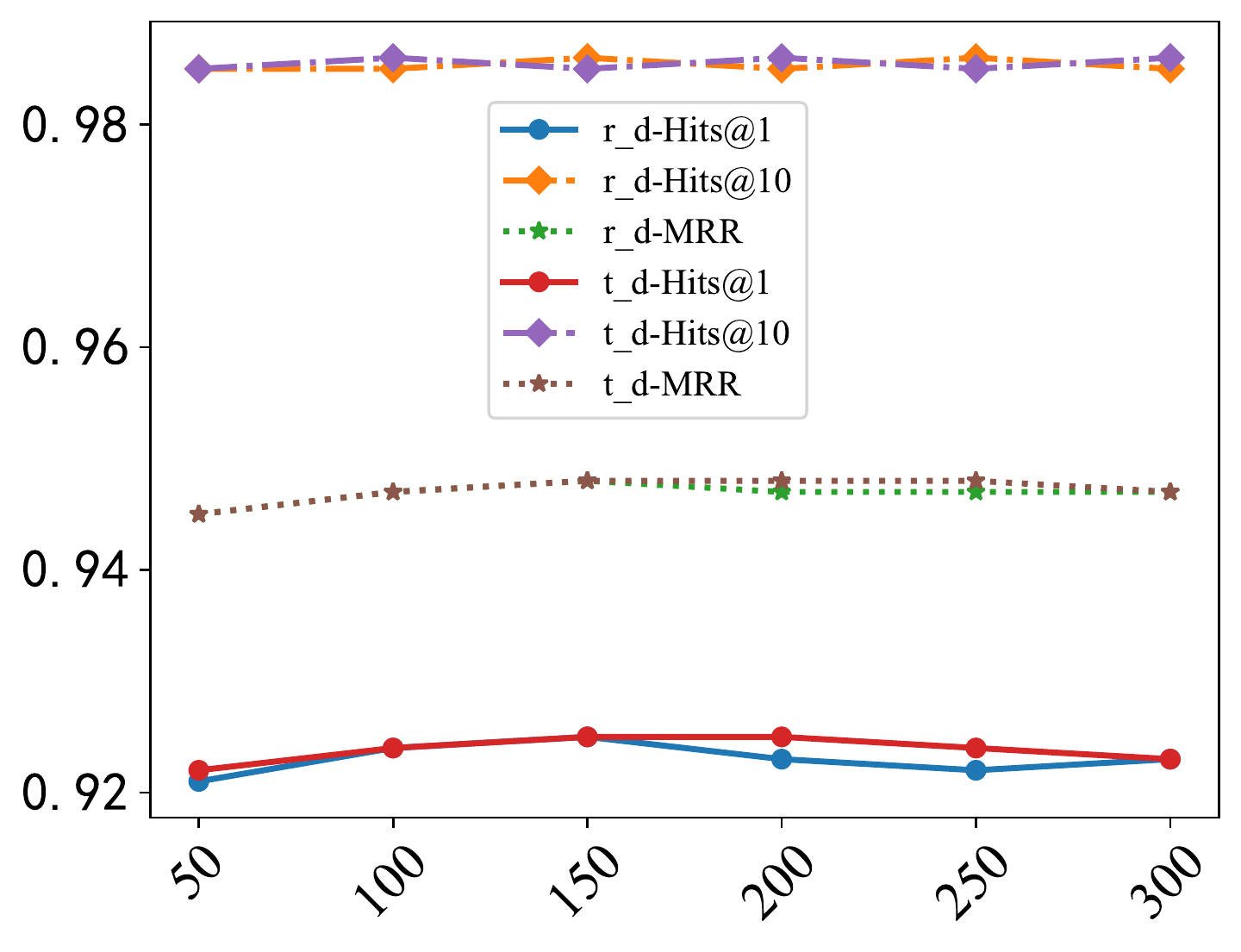}
		\end{minipage}
	}
	\caption{EA performance with different relation and ontology dimensions, \textit{r\_d} denotes the number of relation dimension and \textit{t\_d} is the number of ontology dimension.}
	\label{FIG:dimension}
\end{figure*}
\begin{figure*}[!h]
	\centering
	\subfigure[ZH\_EN]{
		\begin{minipage}{0.3\textwidth}
			\centering
			\includegraphics[width=\textwidth]{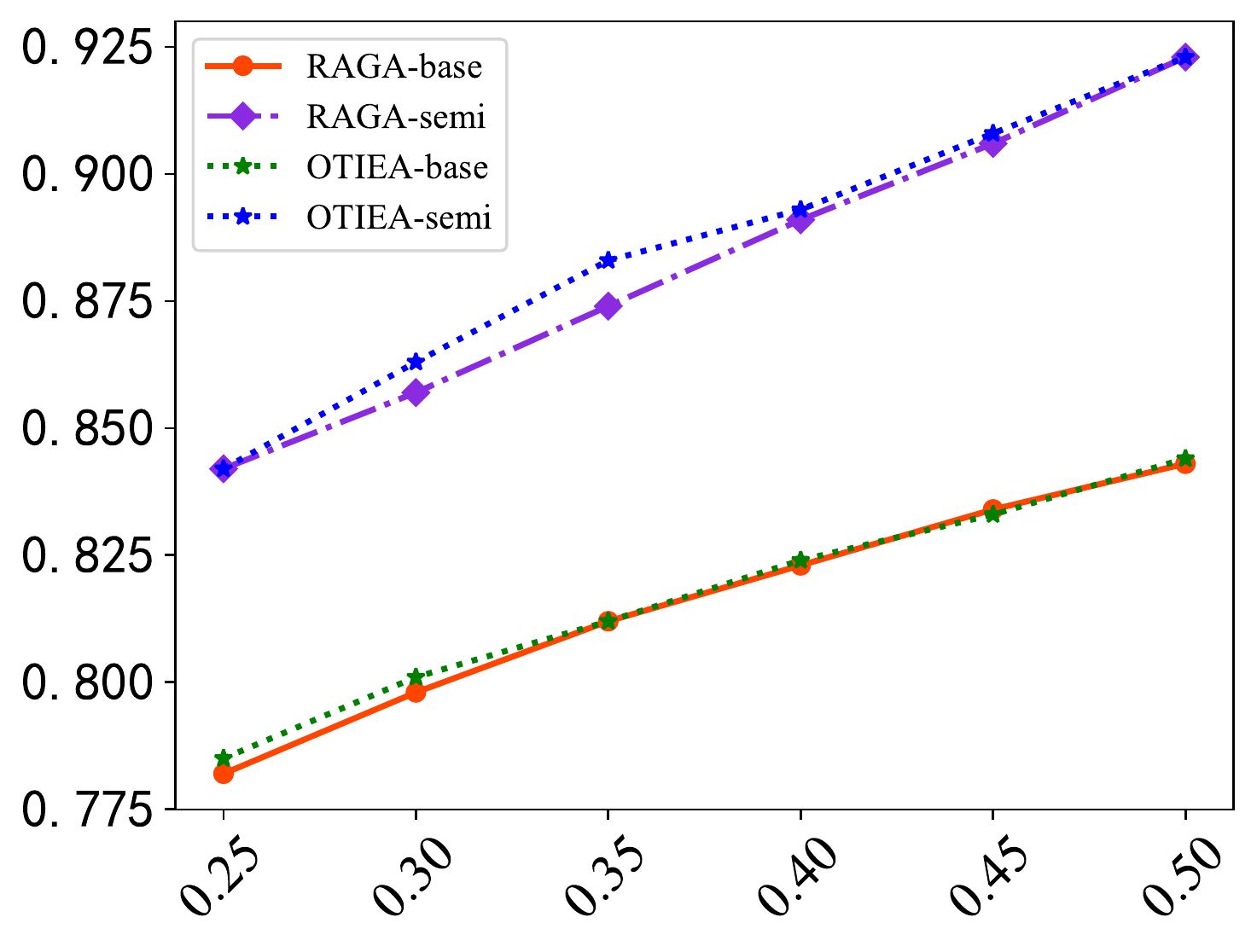}
		\end{minipage}
	}
	\subfigure[JA\_EN]{
		\begin{minipage}{0.3\textwidth}
			\centering
			\includegraphics[width=\textwidth]{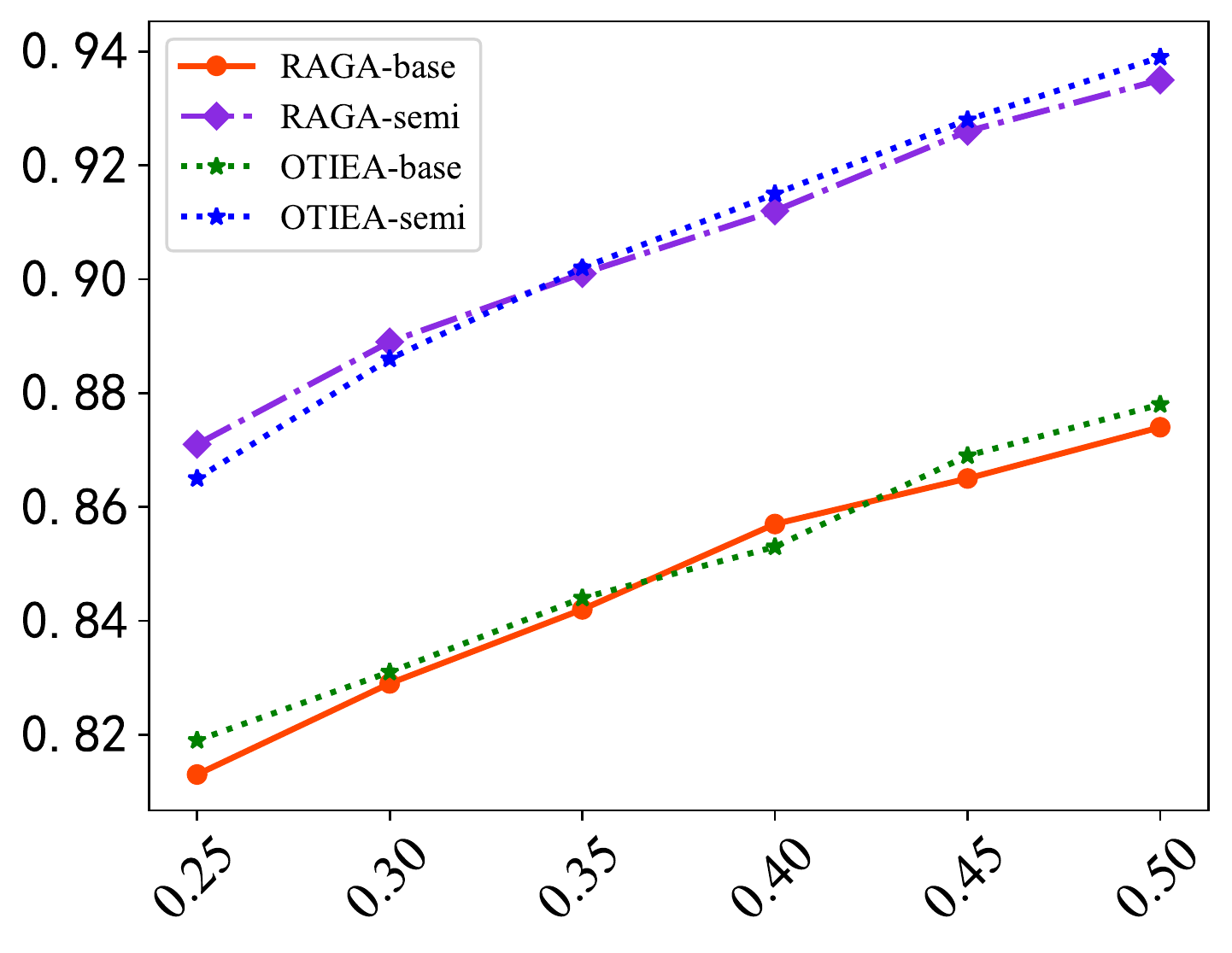}
		\end{minipage}
	}
	\subfigure[FR\_EN]{
		\begin{minipage}{0.3\textwidth}
			\centering
			\includegraphics[width=\textwidth]{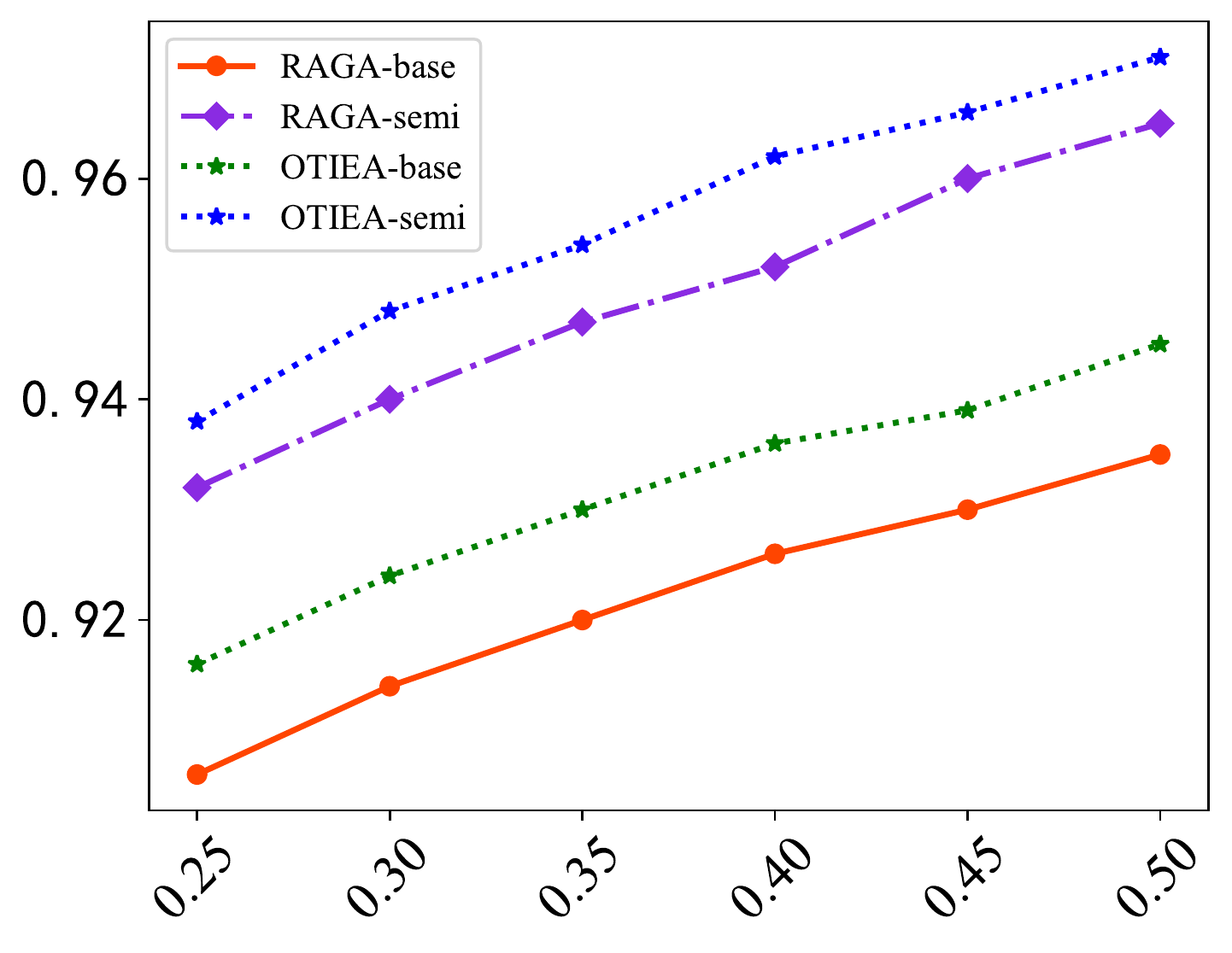}
		\end{minipage}
	}
	\caption{Hits@1 with different ratios of training seed entity pairs.}
	\label{FIG:train1}
\end{figure*}
%#训练集比率不同效果Hits@10对比
\begin{figure*}[!h]
	\centering
	\subfigure[ZH\_EN]{
		\begin{minipage}{0.3\textwidth}
			\centering
			\includegraphics[width=\textwidth]{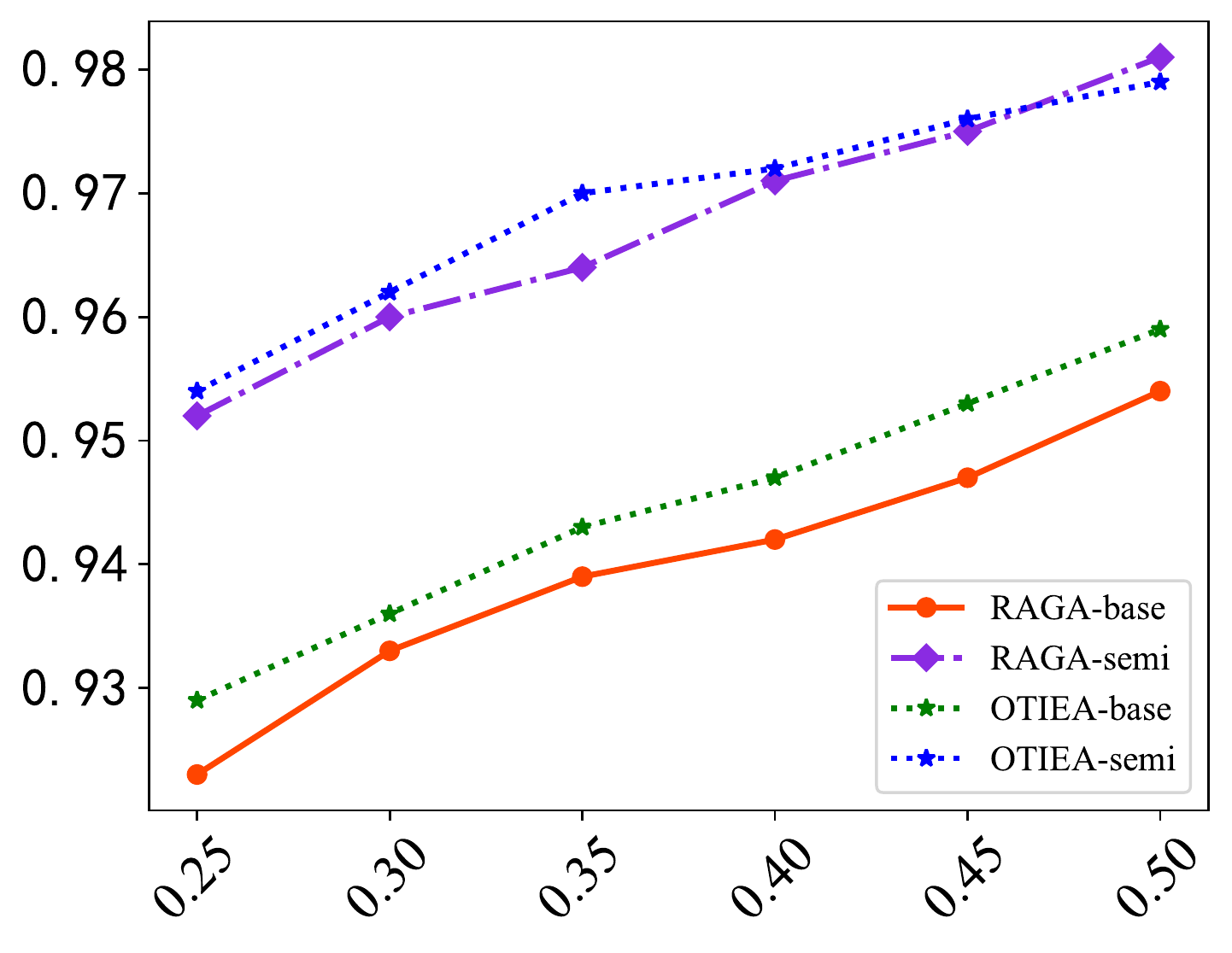}
		\end{minipage}
	}
	\subfigure[JA\_EN]{
		\begin{minipage}{0.3\textwidth}
			\centering
			\includegraphics[width=\textwidth]{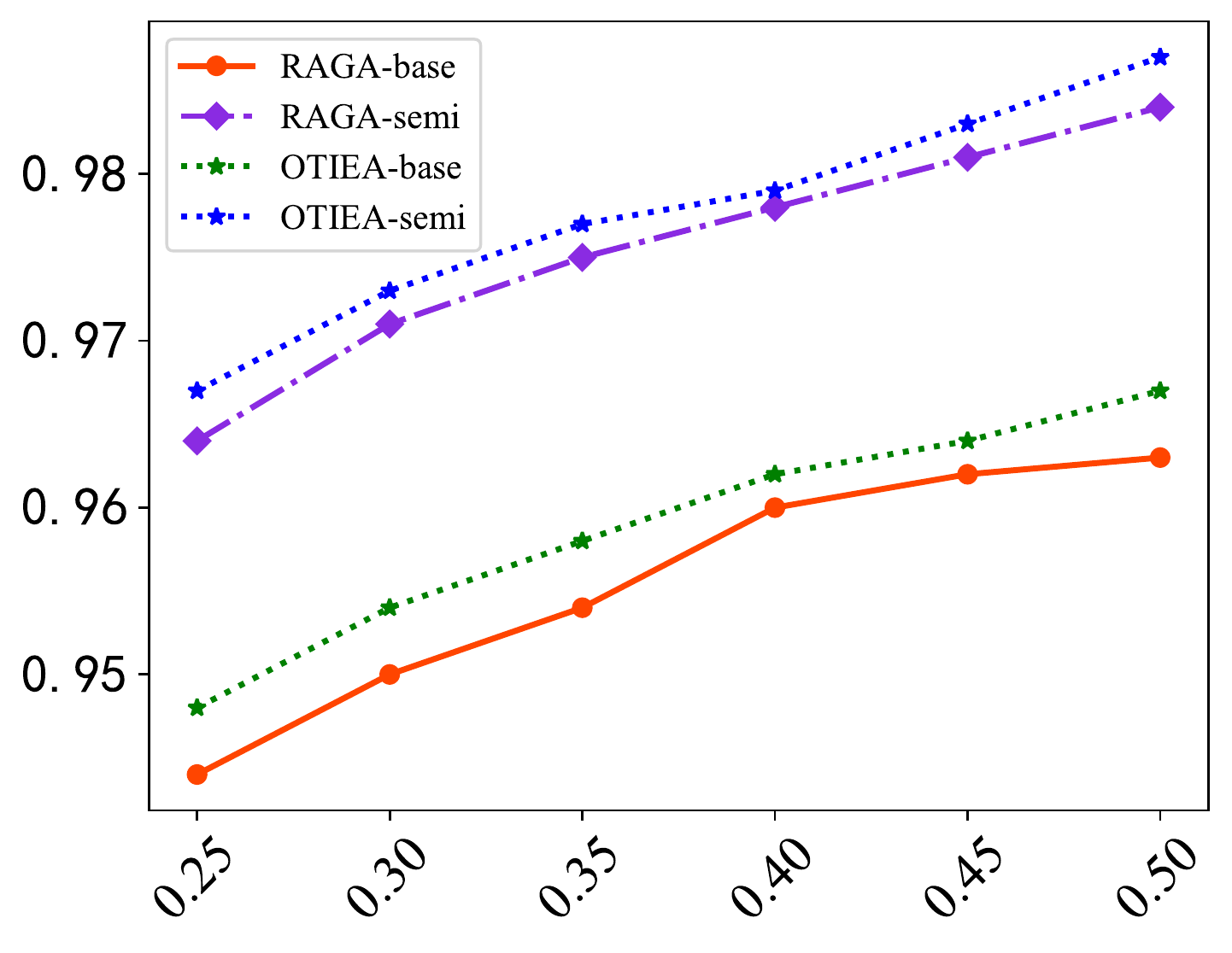}
		\end{minipage}
	}
	\subfigure[FR\_EN]{
		\begin{minipage}{0.3\textwidth}
			\centering
			\includegraphics[width=\textwidth]{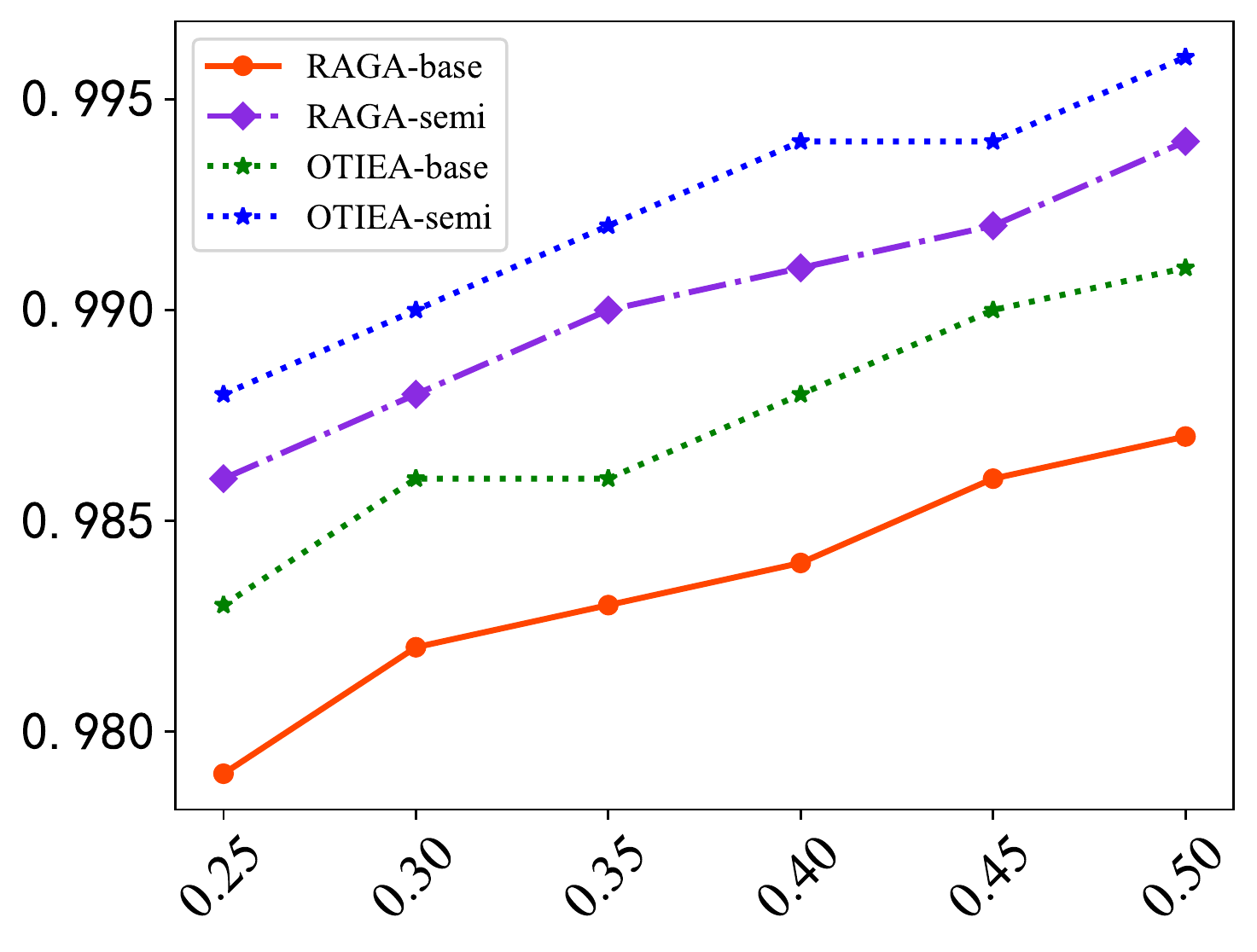}
		\end{minipage}
	}
	\caption{Hits@10 with different ratios of training seed entity pairs.}
	\label{FIG:train10}
\end{figure*}
%#训练集比率不同效果MRR对比
\begin{figure*}[!h]
	\centering
	\subfigure[ZH\_EN]{
		\begin{minipage}{0.3\textwidth}
			\centering
			\includegraphics[width=\textwidth]{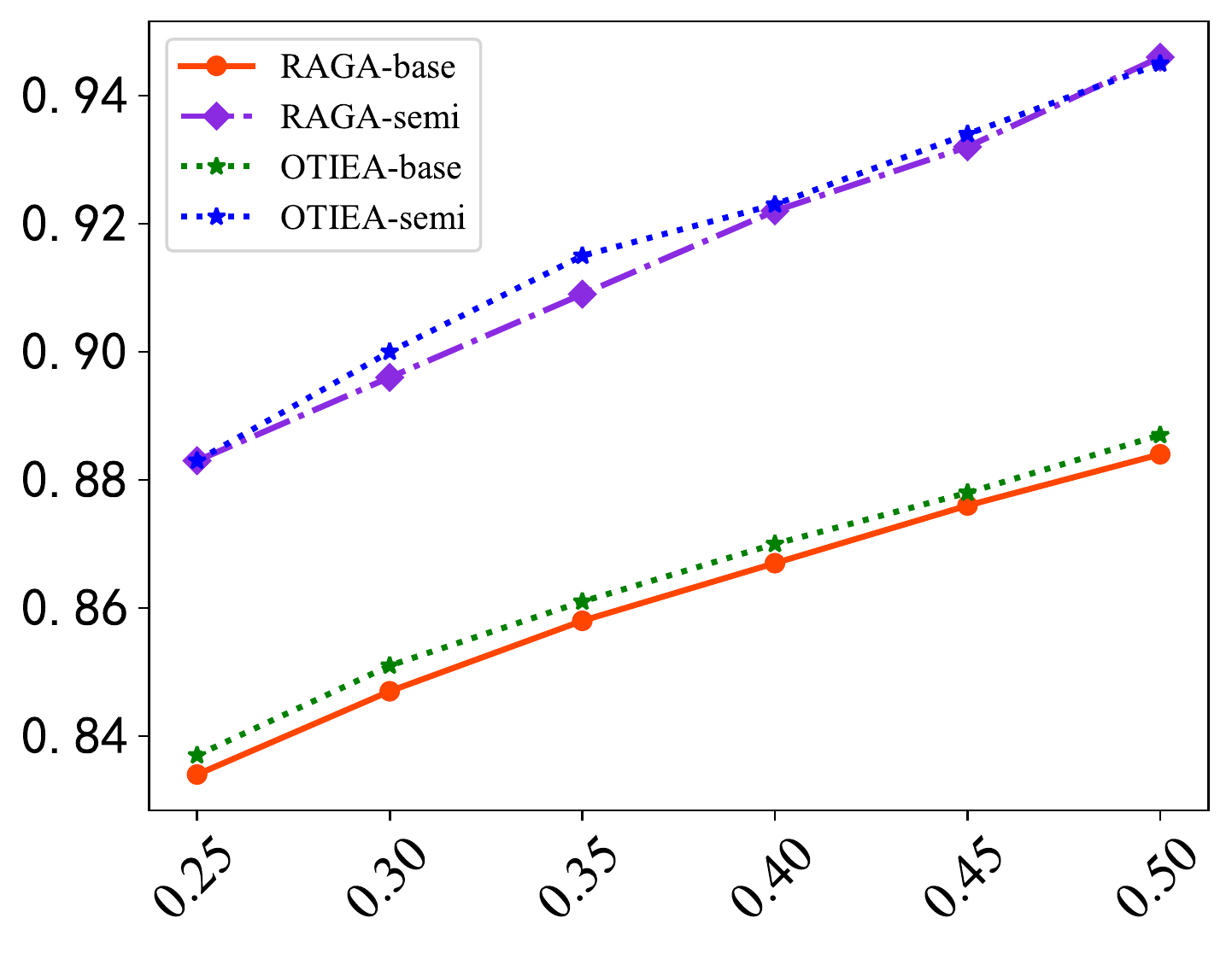}
		\end{minipage}
	}
	\subfigure[JA\_EN]{
		\begin{minipage}{0.3\textwidth}
			\centering
			\includegraphics[width=\textwidth]{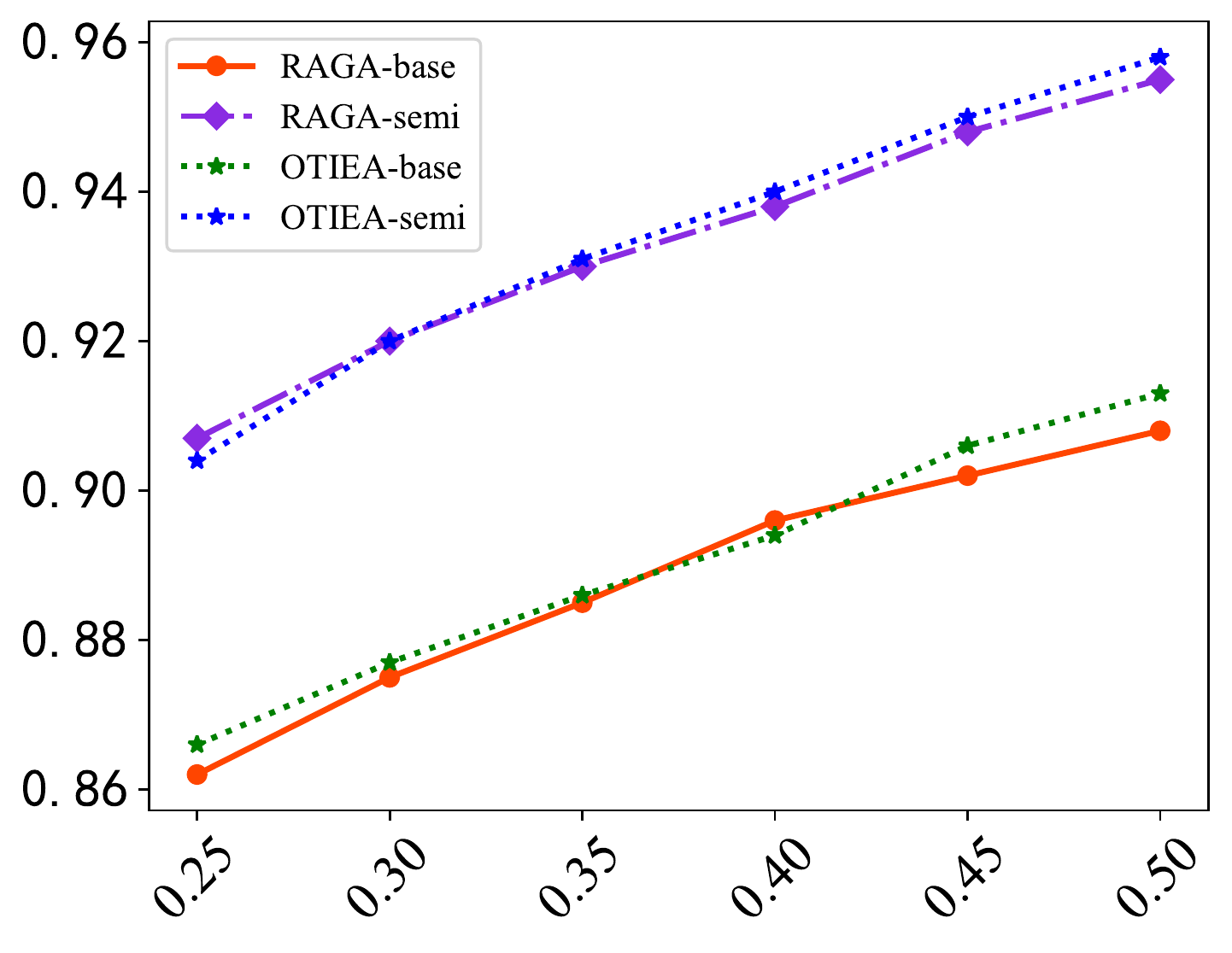}
		\end{minipage}
	}
	\subfigure[FR\_EN]{
		\begin{minipage}{0.3\textwidth}
			\centering
			\includegraphics[width=\textwidth]{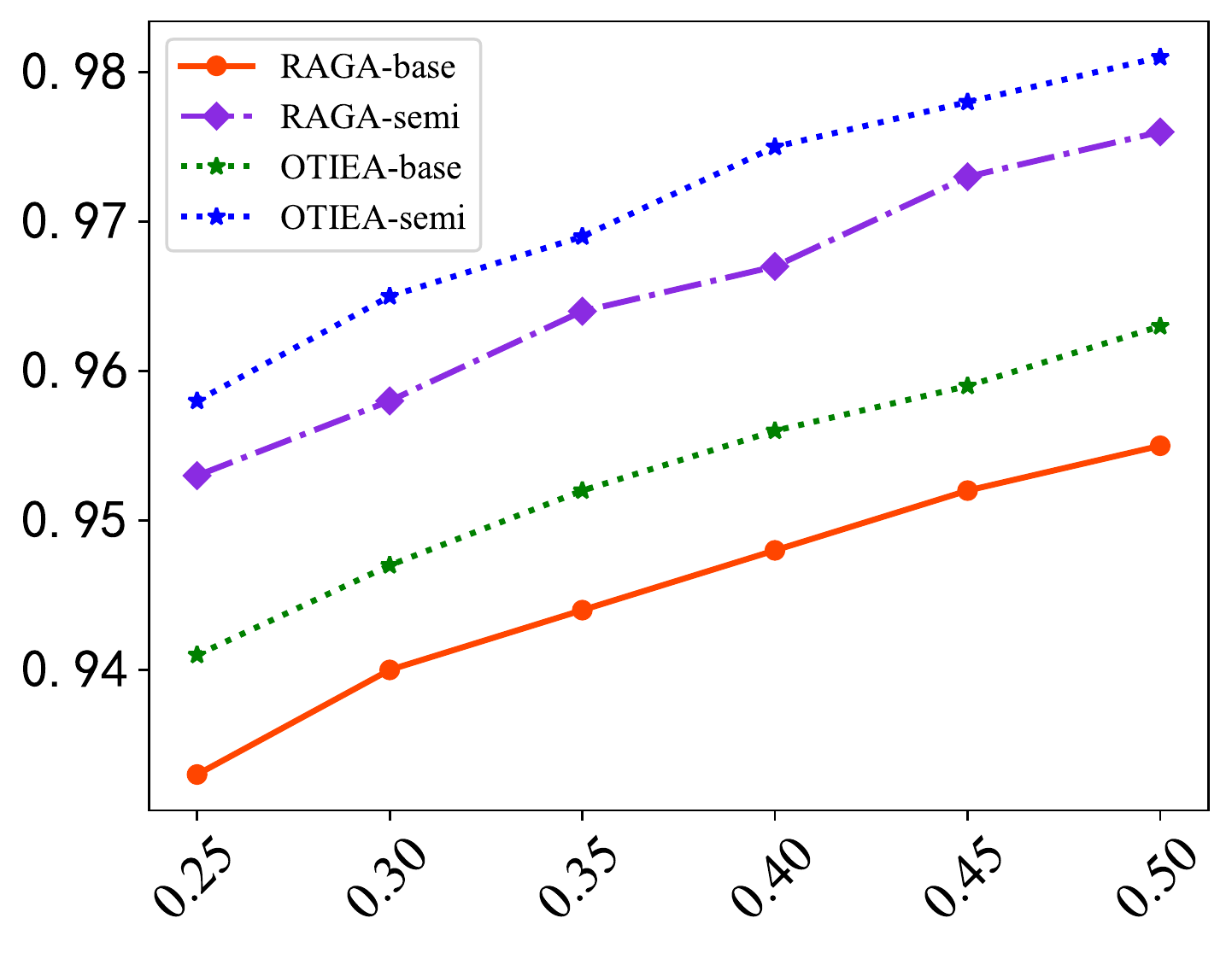}
		\end{minipage}
	}
	\caption{MRR with different ratios of training seed entity pairs.}
	\label{FIG:trainmrr}
\end{figure*}

The comparison of overall experimental results shows that our framework achieves a competitive performance across all three sub-datasets and metrics. Compared with almost all baseline methods that do not utilize external resources, OTIEA performs better in both basic and semi-supervised versions. The possible reason is that OTIEA models triple ensemble information via triple-level attention, which further suggests that the triple should be tackled as a whole rather than loose components in practice. Specifically, Trans-based approaches generally achieve low performance for the lack of representation ability. On the contrary, GNNs-based methods have made great progress in recent years, in which RAGA \citep{zhu_raga_2021} is the most related solution to our work introducing attention mechanism between entity and relation while ignoring the triple ensemble features and the influence of entity type. In detail, OTIEA-base exceeds the state-of-the-art RAGA on Hits@1, Hits@10 and MRR, while OTIEA-semi is slightly worse than RAGA-semi on Hits@1 across JA\_EN, which may be due to the difference of data distribution in data set. For the variants, OTIEA-base obtains a better performance on almost all metrics compared with OTIEA-base(wo-E), OTIEA-base(wo-O) and OTIEA-base(wo-C) proving that the global triple representation, ontology information and cycle co-enhanced mechanism show a certain positive impact on the overall framework.

To further verify the robustness of our framework, we compared OTIEA-base and OTIEA-semi with RAGA-base and RAGA-semi using different ratios of training seed entity pairs on three sub-datasets. Figure \ref{FIG:train1}, Figure \ref{FIG:train10} and Figure \ref{FIG:trainmrr} depict the Hits@1, Hits@10 and MRR results with different seed entity pair ratios varying from 25\% to 50\%, respectively. The results show that OTIEA achieves a significant improvement on FR\_EN across all metrics and obtains a slightly superiority or approximation on ZH\_EN and JA\_EN especially across Hits@10.

To investigate the effectiveness of different GCN depths, we compared three OTIEA-base variants owning different GCN depths as $l=1$, $l=2$ and $l=3$ in Table \ref{gcn}. The results show that the dataset is sensitive to GCN depth, specially, a two-layer GCNs make greatest advantages on ZH\_EN and JA\_EN, while a one-layer GCN is best for FR\_EN in the whole.

To further explore the impact of relation and ontology dimensions, we make ablation studies with different relation and ontology dimensions from 50 to 300. The results of Hits@1, Hits@10 and MRR on different relation and ontology dimension are displayed as Figure \ref{FIG:dimension}. We can see from the dimensional curves that the framework variants with different relation and ontology dimensions have an approximate performance indicating the weak influence of dimension on model. The possible causes are that the embedding dimensions represent the reflection from real space to latent low-dimensional vector space indicating that the dimension number is unimportant as long as the expression requirements are met.

To prove the availability of cycle co-enhanced mechanism, we made a detailed comparison on different modes of cycle co-enhanced mechanism as Figure \ref{FIG:modes}:

\textit{\textbf{mode1}}: the mode of co-enhanced process with the head-tail order and without cycle process.

\textit{\textbf{mode2}}: the mode of cycle co-enhanced process with the head-tail-head order.

\textit{\textbf{mode3}}: the mode of cycle co-enhanced process with the head-tail-head-tail order.

The results in Table \ref{cycle} show that the \textbf{\emph{mode2}} used in our primary OTIEA framework achieved the best performance on three sub-datasets across all metrics, which illustrates the effectiveness of the cycle co-enhance mechanism with \textbf{\emph{mode2}} style.
%\begin{figure}[!h]
%	\centering
%	\subfigure[]{
	%		\begin{minipage}{0.14\textwidth}
		%			\centering
		%			\includegraphics[width=\textwidth]{figs/ZH_EN_xfusion_bar.pdf}
		%		\end{minipage}
	%	}
%	\subfigure[]{
	%		\begin{minipage}{0.14\textwidth}
		%			\centering
		%			\includegraphics[width=\textwidth]{figs/JA_EN_xfusion_bar.pdf}
		%		\end{minipage}
	%	}
%	\subfigure[]{
	%		\begin{minipage}{0.14\textwidth}
		%			\centering
		%			\includegraphics[width=\textwidth]{figs/FR_EN_xfusion_bar.pdf}
		%		\end{minipage}
	%	}
%	\caption{Comparison of different modes of Cycle co-Enhanced module}
%	\label{FIG:cycle}
%\end{figure}

\begin{table}[!h]
	\caption{Comparison of different modes of Cycle Co-enhanced mechanism.\label{cycle}}
	\resizebox{\linewidth}{!}{
		\begin{tabular}{@{\extracolsep{\fill}}lccccccccc@{\extracolsep{\fill}}}
			\toprule%
			& \multicolumn{3}{@{}c@{}}{\textbf{ZH-EN}} & \multicolumn{3}{@{}c@{}}{\textbf{JA-EN}} & \multicolumn{3}{@{}c@{}}{\textbf{FR-EN}} \\\cmidrule{2-4}\cmidrule{5-7} \cmidrule{8-10}%
			\textbf{Modes}& H@1 &H@10 &MRR& H@1 &H@10 &MRR &H@1& H@10 &MRR\\
			\midrule
			\textit{mode1} &79.4 &93.4 &0.846 &81.7 &95.0 &0.866 &91.9 &98.5 &0.944\\
			\textit{mode2} &\textbf{80.1} &\textbf{93.6} &\textbf{0.851} &\textbf{83.1}&\textbf{95.4}&\textbf{0.877}&\textbf{92.4}&\textbf{98.6}&\textbf{0.947}\\
			\textit{mode3} &79.8 &\textbf{93.6} &0.849 &83.0 &95.3 &0.876 &92.3 &98.5 &\textbf{0.947}\\
			\bottomrule
		\end{tabular}
	}
\end{table}
\begin{table}[!h]
	\caption{Comparison of different depths of GCNs.\label{gcn}}
	\resizebox{\linewidth}{!}{
		\begin{tabular}{@{\extracolsep{\fill}}lccccccccc@{\extracolsep{\fill}}}
			\toprule%
			& \multicolumn{3}{@{}c@{}}{\textbf{ZH-EN}} & \multicolumn{3}{@{}c@{}}{\textbf{JA-EN}} & \multicolumn{3}{@{}c@{}}{\textbf{FR-EN}} \\\cmidrule{2-4}\cmidrule{5-7} \cmidrule{8-10}%
			\textbf{Depths}& H@1 &H@10 &MRR& H@1 &H@10 &MRR &H@1& H@10 &MRR\\
			\midrule
			\textit{l}=1 &79.2 &92.1 &0.840 &\textbf{83.6} &94.6 &0.876 &\textbf{93.2} &98.5 &\textbf{0.952}\\
			\textit{l}=2 &\textbf{80.1} &\textbf{93.6} &\textbf{0.851} &83.1 &\textbf{95.4} &\textbf{0.877} &92.4 &\textbf{98.6} &0.947\\
			\textit{l}=3 &76.8 &\textbf{93.5} &0.830 &78.8 &94.4 &0.846 &88.4 &97.5 &0.919\\
			\bottomrule
		\end{tabular}
	}
\end{table}
%\begin{figure}[!h]
%	\centering
%	\subfigure[]{
	%		\begin{minipage}{0.12\textwidth}
		%			\centering
		%			\includegraphics[width=\textwidth]{figs/ZH_EN_gcn_bar.pdf}
		%		\end{minipage}
	%	}
%	\subfigure[]{
	%		\begin{minipage}{0.12\textwidth}
		%			\centering
		%			\includegraphics[width=\textwidth]{figs/JA_EN_gcn_bar.pdf}
		%		\end{minipage}
	%	}
%	\subfigure[]{
	%		\begin{minipage}{0.12\textwidth}
		%			\centering
		%			\includegraphics[width=\textwidth]{figs/FR_EN_gcn_bar.pdf}
		%		\end{minipage}
	%	}
%	\caption{Comparison of different depths of Highway-GCNs}
%	\label{FIG:gcn}
%\end{figure}

\section{Conclusion}\label{sec7}
In this paper, a novel multi-level attention-based entity alignment model OTIEA is proposed. It adopts an end-to-end framework to learn representative entity embeddings for EA. The ontology-enhanced triple encoder generates ensemble triple representation via a three-stage internal interaction attention mechanism considering both intrinsic correlation of triple and ontology pair enhancement, and the triple-aware entity decoder circularly propagates the triple information to EA-orient entity representations according to roles features via triple-aware attention. Furthermore, The encoder-decoder architecture is capable to tackle overlapping problem naturally by considering triple indivisibility and regarding entity pair as a whole in ontology enhancement stage. The experiments on three real-world datasets and further ablation studies demonstrate the effectiveness of our framework.
\bmhead{Acknowledgements} This work is supported by the National Natural Science Foundation of China under Grant NO.72274138.
\section*{Statements and Declarations}
\begin{itemize}
	\item \textbf{Conflict of interest/Competing interests:} The authors have no relevant financial or non-financial interests to disclose.
	\item \textbf{Consent to participate:} All authors declare the consent to participate.
	\item \textbf{Consent for publication:} All authors declare the consent for publication.
	\item \textbf{Authors' contributions:} All authors contributed to the study conception and design. Material preparation, data collection and analysis were performed by Chengxiang Tan, Zhishuo Zhang. Article modification was executed by Xueyan Zhao, Min Yang and Chaoqun Jiang. The first draft of the manuscript was written by Zhishuo Zhang and all authors commented on previous versions of the manuscript. All authors read and approved the final manuscript.
	\item \textbf{Data Availability and Access:}
	The data used in this paper is open source on the network, and we provide the data in our available online resources in https://github.com/CodesForNlp/OTIEA.
	\item \textbf{Compliance with Ethical Standards:} The authors declare the informed consent and there is no ethics involved.
\end{itemize}


\bibliographystyle{bst/sn-standardnature}
\bibliography{Alignment}

\end{CJK}
\end{sloppypar}
\end{document}